\documentclass[letterpaper]{article} 
\usepackage[preprint]{aaai2027}  
\usepackage[hyphens]{url}  
\usepackage{graphicx} 
\usepackage{natbib}  
\usepackage{caption} 
\usepackage{algorithm}

\usepackage{newfloat}
\usepackage{listings}
\DeclareCaptionStyle{ruled}{labelfont=normalfont,labelsep=colon,strut=off} 
\floatstyle{ruled}
\newfloat{listing}{tb}{lst}{}
\floatname{listing}{Listing}

\usepackage{booktabs}

\usepackage{float}
\usepackage{algpseudocode}
\usepackage{amsmath}
\usepackage{amssymb}
\usepackage{multirow} 
\usepackage{subcaption}
\usepackage{blindtext}

\newcommand{\bx}{\boldsymbol{x}}

\newcommand{\bu}{\boldsymbol{u}}

\newcommand{\cG}{\mathcal{G}}

\title{MGE: A Membership Guided Data Extraction Attack against Diffusion Models}
\author{
    Yuchang Mei,
    Longqiang Wang,
    Haoyang Jiang,
    Mingyang Yi\textsuperscript{*},\\
    Shaolei Zhang,
    Keman Huang,
    Wuqiong Pan,
    Ju Fan
}
\affiliations{
    Renmin University of China, Beijing, China\\
    \texttt{meiziqing10@outlook.com}
}

\makeatletter
\g@addto@macro\@maketitle{%
    \begin{minipage}{\textwidth}
        \centering
        \includegraphics[
            width=\textwidth,
            height=\textheight,
            keepaspectratio
        ]{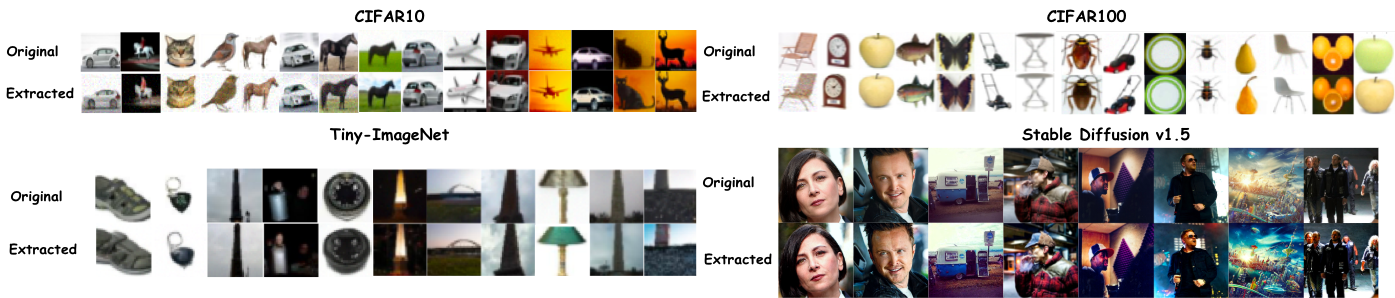}
        \captionof{figure}{
            Examples of images extracted from pretrained diffusion models by our proposed MGE on different datasets.
        }
        \label{fig:extracted}
    \end{minipage}
}
\makeatother

\begin{document}

\maketitle

\begin{abstract}
Recently, diffusion models have achieved remarkable performance in image generation. However, their inherent memory properties introduce a potential safety risk of training data leakage, i.e., data extraction attack. To illustrate such risk, we propose a membership guided extraction (MGE) framework to effectively extract training data from a pre-trained diffusion model. Our core idea is to formulate the data extraction as a conditional generation, where the condition signal is from a membership inference classifier. To improve the success rate and efficiency of MGE, we propose adversarial gradient and parallel branch decoding. Our experiments on image generation benchmarks demonstrate that our MGE can extract hundreds of images, which shows that more attention should be paid to privacy-preserving of diffusion model.
\end{abstract}


\section{Introduction}
\label{sec:intro}

\begin{figure*}[h]
    \centering
    \includegraphics[width=\textwidth]{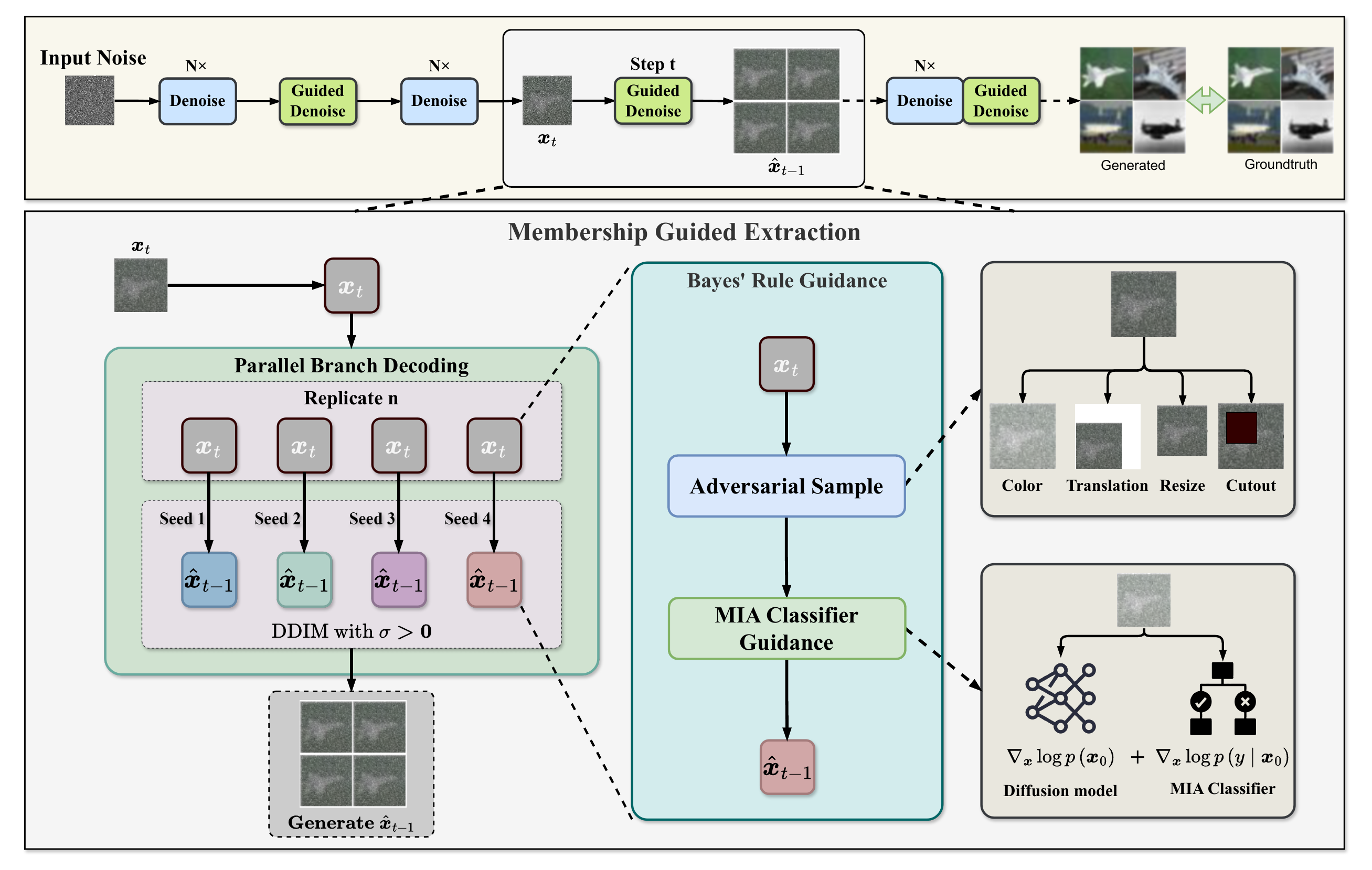}
    \caption{An overview of our proposed Membership Guidance Extraction MGE. MGE, which is based on Bayes' rule, consists of two main modules: adversarial guidance and parallel branch decoding. It aims to transform the data extraction attack into a conditional generation process with signals constructed by a membership inference attacker.}
    \label{fig:MGE_algo}
\end{figure*}

Recently, the diffusion probabilistic model \cite{ho2020denoising, song2019generative, song2021scorebased} has emerged as a powerful paradigm in generative tasks. It has achieved many state-of-the-art results in a wide variety of practical applications, including image generation \cite{rombach2022high, saharia2022photorealistic}, audio synthesis \cite{popov2021grad, kong2020diffwave}, video generation \cite{ho2022video, yang2023diffusion}, and many other important domains \cite{rasul2021autoregressive, hoogeboom2022equivariant, li2022diffusion}. 

Though diffusion models have good generation capabilities, they have potential privacy risks as discussed in \cite{carlini2023extracting}. The existing risks are mainly revealed by two attacks arising from the inherent memorization capabilities of diffusion models \cite{wen2024detecting,carlini2023extracting}. The membership inference attack \cite{hu2023membership, kong2023efficient, duan2023diffusion} and data extraction attack, which respectively detect and extract data from the training set.

In this paper, we focus on the data extraction attack of diffusion models \cite{carlini2023extracting, webster2023reproducible}. Early diffusion-model data extraction attacks largely followed the ``generate-and-filter'' paradigm \cite{carlini2023extracting,nasr2023scalable,carlini2021extracting}, in which generated candidates are screened for memorized training examples. More recent active extraction methods, including SIDE \cite{chen2026side} and FineXtract \cite{wu2025finextract}, also explore guided extraction, but use different guidance signals and attack settings. In this paper, we instead construct guidance directly from training-set membership scores derived from the target diffusion model, without training an additional auxiliary classifier. Specifically, we use classifier-guided generation (CG) \cite{song2021scorebased,yi2024towards} to direct sampling toward the desired membership condition. To conduct data extraction attack, a natural idea is to set the condition as ``\emph{from the training set}''. Clearly, this condition can be realized by a classifier constructed by MIA \cite{kong2023efficient}, which predicts whether a sample is from the training set or not. Our classifier acts as a surrogate of MIA attacker \cite{kong2023efficient}, which predicts whether a generated sample belongs to the training set. Then, our MIA classifier intuitively guides the model to generate data from target class (in training set). 

Concretely, we formulate the data extraction attack on diffusion models as a CG, where the condition is a designed membership inference classifier making the predictions based on a ``reconstruction error'' \cite{song2020denoising,kong2023efficient}. However, naively incorporating this classifier into the generation process has two limitations. Firstly, the ``adversarial samples'' \cite{goodfellow2014explaining,arrieta2020explainable,liu2024referee,dinh2023rethinking} may deceive the classifier and generate samples that are not in the training set as desired. To resolve this, we propose using adversarial gradient to construct guidance signal as in \cite{papernot2016transferability, feinman2017detecting, shen2024understanding}. Secondly, injecting the guidance signal slows down the generation process. Thus, we propose injecting the guidance signal dynamically and using random parallel decoding techniques to accelerate the generation process.  

All these technical components together consist of our Membership Guidance Extraction (MGE) to extract training data from the diffusion model as shown in Figure \ref{fig:MGE_algo}. Empirically, we apply our MGE to Stable Diffusion v1.5 \cite{rombach2022high} and other models trained on benchmarks \texttt{CIFAR10}, \texttt{CIFAR100} \cite{krizhevsky2009learning}, and \texttt{TinyImageNet} \cite{deng2009imagenet}. The results indicate that our MGE successfully extracts hundreds of images within $2^{17}$ generations. Besides that, our MGE improves the existing baseline \cite{carlini2023extracting} in both success rate and efficiency.

\section{Related Works}
\label{sec:related}
Diffusion probabilistic models \cite{ho2020denoising,song2020denoising} have demonstrated their powerful ability in image generation. However, \cite{carlini2023extracting,wang2025improved} argued that diffusion models are more vulnerable to privacy attack \cite{carlini2023extracting, somepalli2023diffusion,duan2023diffusion}, compared to other generative models such as GANs \cite{goodfellow2014explaining}, VAEs \cite{kingma2013auto,ramesh2021zero}. 

\paragraph{Membership Inference Attack.}  Membership Inference Attack (MIA) \cite{shokri2017membership, carlini2022membership, truong2025attacks} aims to detect whether a given sample belongs to the training set of a target model, with the model available (white-box attack) \cite{pang2023white, duan2023diffusion} or not (black-box attack) \cite{tang2023membership, pang2023black, li2024towards, kong2023efficient}. First deployed on classification model \cite{shokri2017membership, salem2018ml, choquette2021label, ye2022enhanced}, MIA has been generalized to generative models, including large language models (LLMs) \cite{carlini2021extracting} and diffusion models \cite{kong2023efficient,duan2024membership}. In this paper, we focus on diffusion models, whereas the MIA are constructed using either ``reconstruction error'' \cite{duan2023diffusion,duan2024membership} or ``noise difference'' \cite{kong2023efficient}. These features are exploited to construct classifiers for membership inference \cite{carlini2022membership, carlini2023extracting, somepalli2023diffusion}. Moreover, the MIA is developed to some specific scenarios including Text-to-Image generation \cite{li2024shake}, DreamBooth \cite{li2024shake}, and LoRA \cite{luo2025privacy}. In this paper, we use the MIA from \cite{kong2023efficient} to construct the signal that guides model to generate data from the training set.   

\paragraph{Data Extraction Attack.} Data Extraction Attack (DEA) refers to an attacker reconstructing examples from the training set through queries or other access methods to the model \cite{carlini2021extracting,somepalli2023diffusion}. Similar to MIA, the DEA was first developed on classification models \cite{fredrikson2015model,he2019model, zhang2020secret}, and then generalized to generative models. For example, Carlini et al. \cite{carlini2021extracting} recover sensitive training text samples (e.g., keys, personal information, or complete document fragments) through constructed prompts/queries from autoregressive language models \cite{devlin2019bert}. The technique was then generalized to LLMs by \cite{nasr2023scalable}, including LLaMA \cite{touvron2023llama}, ChatGPT \cite{brown2020language}, and Falcon \cite{almazrouei2023falcon}. In this paper, we focus on the DEA on image diffusion models. Unlike generate-and-filter extraction \cite{carlini2023extracting}, our attack incorporates a membership-based guidance signal directly into generation. Other guided approaches differ in how this signal is constructed: SIDE \cite{chen2026side} uses cluster-based surrogate conditions that do not explicitly encode training-set membership, while FineXtract \cite{wu2025finextract} leverages the differences between pretrained and fine-tuned model checkpoints to extract fine-tuning data. MGE instead derives its guidance from the target model itself, without training a separate auxiliary classifier or requiring a pretrained--fine-tuned checkpoint pair. 

\section{Preliminary}
\label{sec:preliminary}

\subsection{Classifier Guided Diffusion Model}
The Denoising Diffusion Probabilistic Model (DDPM) \cite{ho2020denoising,song2021scorebased} is a powerful probabilistic generative model that generates the target through a gradual denoising process. Technically, the original data $\boldsymbol{x}_{0}\sim p_{0}(\boldsymbol{x}_{0})$ is interrupted as $\boldsymbol{x}_{t} = \sqrt{\bar{\alpha}_{t}}\boldsymbol{x}_{0} + \sqrt{1 - \bar{\alpha}_{t}}\boldsymbol{\epsilon}_{0}$ where $\bar{\alpha}_{t}\to 0$ and $\boldsymbol{\epsilon}\sim \mathcal{N}(0, \boldsymbol{I})$. Then, a parameterized model $\boldsymbol{\epsilon}_{\theta}$ is trained to predict the noise on $\boldsymbol{x}_{t}$ as 
\begin{equation}
\min_{\theta}\mathbb{E}\left[\left\|\boldsymbol{\epsilon}_{\theta}(\boldsymbol{x}_{t}, t) - \boldsymbol{\epsilon}\right\|^{2}\right]
\end{equation}
With this model, generation can be implemented as a DDIM \cite{song2020denoising} denoising process, starting from a Gaussian (since $\boldsymbol{x}_{T}\approx \mathcal{N}(0, \boldsymbol{I})$)
\begin{equation}
    \begin{aligned}
    \boldsymbol{x}_{t-1} &=
    \sqrt{\bar{\alpha}_{t-1}}
    \left(
        \frac{
            \boldsymbol{x}_t
            - \sqrt{1 - \bar{\alpha}_{t}}\, 
            \boldsymbol{\epsilon}_\theta(\boldsymbol{x}_t, t)
        }{
            \sqrt{\bar{\alpha}_{t}}
        }
    \right) \\
    &\quad + 
    \sqrt{1 - \bar{\alpha}_{t-1} - \sigma^{2}_{t}} \,
    \boldsymbol{\epsilon}_\theta(\boldsymbol{x}_t, t) + \sigma_{t}\boldsymbol{\epsilon}_{t} \\
    & = \mathrm{DDIM}(t, \boldsymbol{x}_{t}, \boldsymbol{\epsilon}_{\boldsymbol{\theta}}, \sigma_{t}),
    \end{aligned}
    \label{eq:ddim_denoising}
\end{equation}
where $\sigma_{t}$ is a noise schedule, and $\boldsymbol{\epsilon}_{t}\sim\mathcal{N}(0, \boldsymbol{I})$. Notably, the generation process can be extended to conditional generation, i.e., sampling from the conditional distribution $p(\boldsymbol{x}_{0} \mid \mathbf{c})$ for some condition $\mathbf{c}$. Since $\boldsymbol{\epsilon}_{\theta}(\boldsymbol{x}_{t}, t) \approx -\sqrt{1 - \bar{\alpha}_{t}}\nabla_{\boldsymbol{x}_{t}}\log p_{t}(\boldsymbol{x}_{t})$ (Tweedie's lemma \cite{efron2011tweedie}), by Bayes' rule and replacing the noise prediction $\boldsymbol{\epsilon}_{\theta}(\boldsymbol{x}_{t}, t)$ in \eqref{eq:ddim_denoising} with 
\begin{equation}
\begin{aligned}
    & -\sqrt{1 - \bar{\alpha}_{t}}\nabla_{\boldsymbol{x}_{t}}\log{p_{t}}(\boldsymbol{x}_{t} \mid \mathbf{c}) \\
    & = -\sqrt{1 - \bar{\alpha}_{t}}\nabla_{\boldsymbol{x}_{t}}\log{p_{t}(\boldsymbol{x}_{t})} -\sqrt{1 - \bar{\alpha}_{t}}\nabla_{\boldsymbol{x}_{t}}\log{p(\mathbf{c} \mid \boldsymbol{x}_{t})} \\
    & \approx \boldsymbol{\epsilon}_{\theta}(\boldsymbol{x}_{t}, t) - \sqrt{1 - \bar{\alpha}_{t}}\nabla_{\boldsymbol{x}_{t}}\log p(\mathbf{c} \mid \boldsymbol{x}_{t}),
\end{aligned} 
\label{eq:bayes_rule}
\end{equation}
we obtain samples from the conditional distribution $p(\boldsymbol{x}_{0}\mid \mathbf{c})$, i.e., the classifier-guidance generation under $p(\mathbf{c}\mid \boldsymbol{x}_{0})$. We refer readers to \cite{ho2022classifier} for more details.  

\subsection{Membership Inference Attacks}
Given a model, MIA constructs a classification rule/model to predict whether a given data point is from its training set. The MIA has been adapted to diffusion models, \cite{kong2023efficient,duan2024membership} whereas the core idea is that data in the training set will correspond to smaller training loss. In this paper, we use the Proximal Initialization Attack (PIA) \cite{kong2023efficient} to construct our guidance signal, since it takes a good balance between efficiency and accuracy. To determine whether $\boldsymbol{x}_{0}$ is from the training set of model $\boldsymbol{\epsilon}_{\boldsymbol{\theta}}$, the PIA constructs the classifier 
\begin{equation}
    f(\boldsymbol{x}_{0}) = \textbf{1}[R_{t, p}(\boldsymbol{x}_{0}) < \tau],
    \label{eq:attack_classify}
\end{equation}
where the reconstruction error $R_{t, p}$ is defined as 
\begin{equation}
\small
    R_{t, p}(\boldsymbol{x}_{0}) = \left\|\boldsymbol{\epsilon}_{\boldsymbol{\theta}}(\boldsymbol{x}_{0}, 0) - \boldsymbol{\epsilon}_{\boldsymbol{\theta}}(\sqrt{\bar{\alpha}_{t}}\boldsymbol{x}_{0} + \sqrt{1 - \bar{\alpha_{t}}}\boldsymbol{\epsilon}_{\boldsymbol{\theta}}(\boldsymbol{x}_{0}, 0), t)\right\|_{p}.
    \label{eq:reconstruction_error}
\end{equation}
Here, $\|\cdot\|_{p}$ is the $\ell_{p}$ norm, $\textbf{1}(\cdot)$ is the indicator function, and $\tau$ is a threshold. Intuitively, PIA predicts $\boldsymbol{x}_{0}$ is from training set when the ``loss'' on $\boldsymbol{x}_{0}$ is small.

\section{Methodology}
\label{sec:method}

In this section, we propose our Membership Guided Extraction (MGE) method to extract data from the diffusion model. Our idea is to transform the data extraction attack into a conditional generation process, with condition signals constructed using the MIA discussed in Section \ref{sec:preliminary}. Furthermore, we design adversarial guidance and parallel branch decoding, which respectively enhance the strength of guidance signals and the efficiency of the attacker.   

\subsection{Data Extraction Attack is a Conditional Generation}
As discussed in Section \ref{sec:intro}, the goal of the data extraction attack (DEA) is to extract data from the training set of a diffusion model. Technically, the DEA can be formulated as a conditional generation $p(\boldsymbol{x}_{0}\mid \boldsymbol{x}_{0}\in \boldsymbol{S})$, where $\boldsymbol{S}$ is the training set. As discussed in Section \ref{sec:preliminary}, the conditional generational process can be implemented when we construct the classifier-guided noise prediction \eqref{eq:bayes_rule} as 
\begin{equation}
    \hat{\boldsymbol{\epsilon}}_{\boldsymbol{\theta}}(\boldsymbol{x}_{t}, t) = \boldsymbol{\epsilon}_{\boldsymbol{\theta}}(\boldsymbol{x}_{t}, t) - \sqrt{1 - \bar{\alpha}_{t}}\nabla_{\boldsymbol{x}_{t}}\log{p(\boldsymbol{x}_{0}\in \boldsymbol{S} \mid \boldsymbol{x}_{t})}.
    \label{eq:guided_noise}
\end{equation}
This follows from the Bayes decomposition $\nabla_{\boldsymbol{x}_t}\log p_t(\boldsymbol{x}_t\mid c)=\nabla_{\boldsymbol{x}_t}\log p_t(\boldsymbol{x}_t)+\nabla_{\boldsymbol{x}_t}\log p(c\mid\boldsymbol{x}_t)$, with $c$ denoting the event that the clean sample belongs to the training set, together with the score--noise-prediction relationship. To obtain a usable conditional generation rule, we therefore need to approximate $p(\boldsymbol{x}_{0}\in\boldsymbol{S}\mid\boldsymbol{x}_{t})$. 

\begin{figure}[t!]
    \centering
    \begin{subfigure}[t]{\linewidth}
        \centering
        \includegraphics[width=0.8\linewidth]{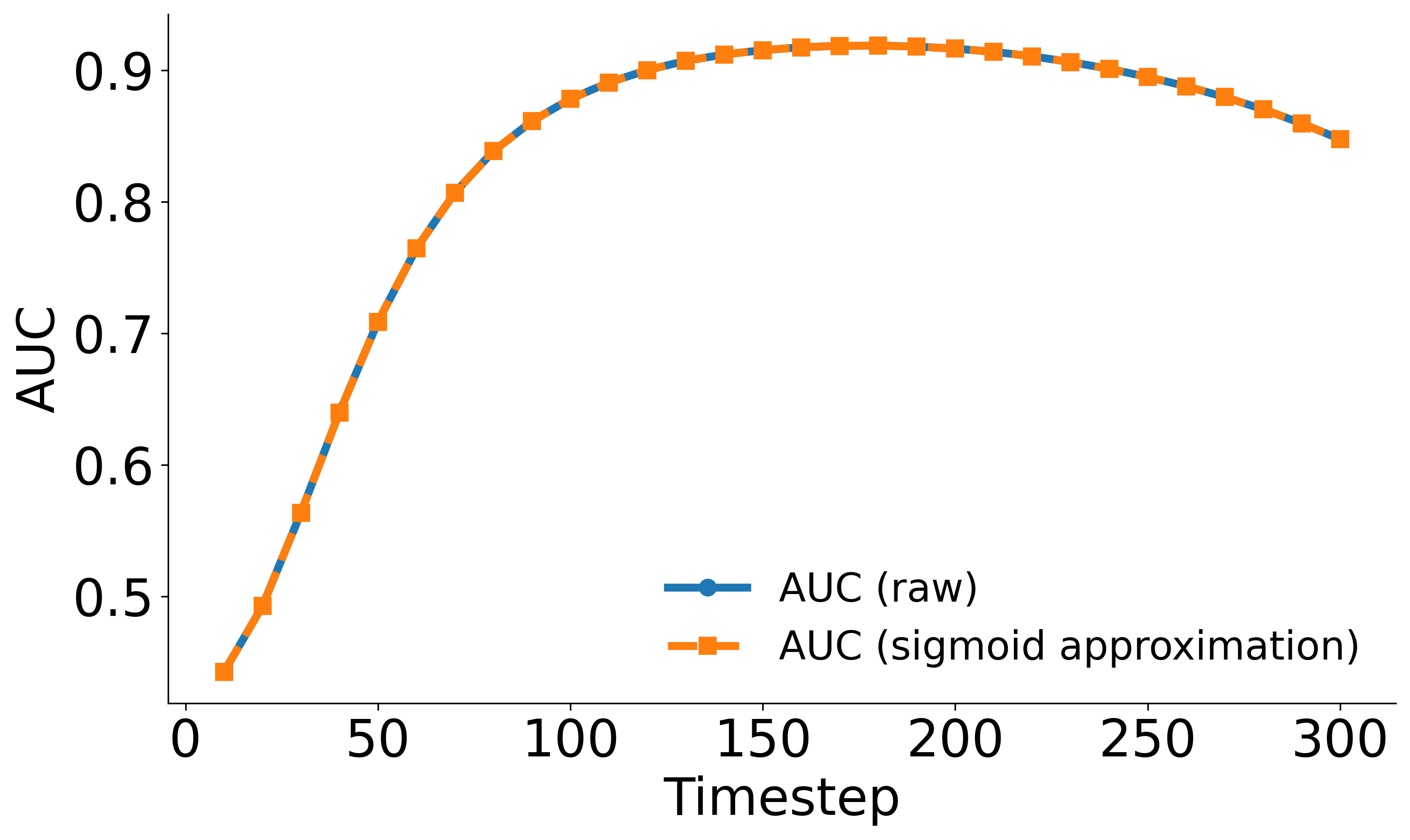}
        \caption{Comparison of AUC under the original membership inference classifier and our approximation.}
        \label{fig:auc_results}
    \end{subfigure}
    \hfill
    \begin{subfigure}[t]{\linewidth}
        \centering
        \includegraphics[width=0.8\linewidth]{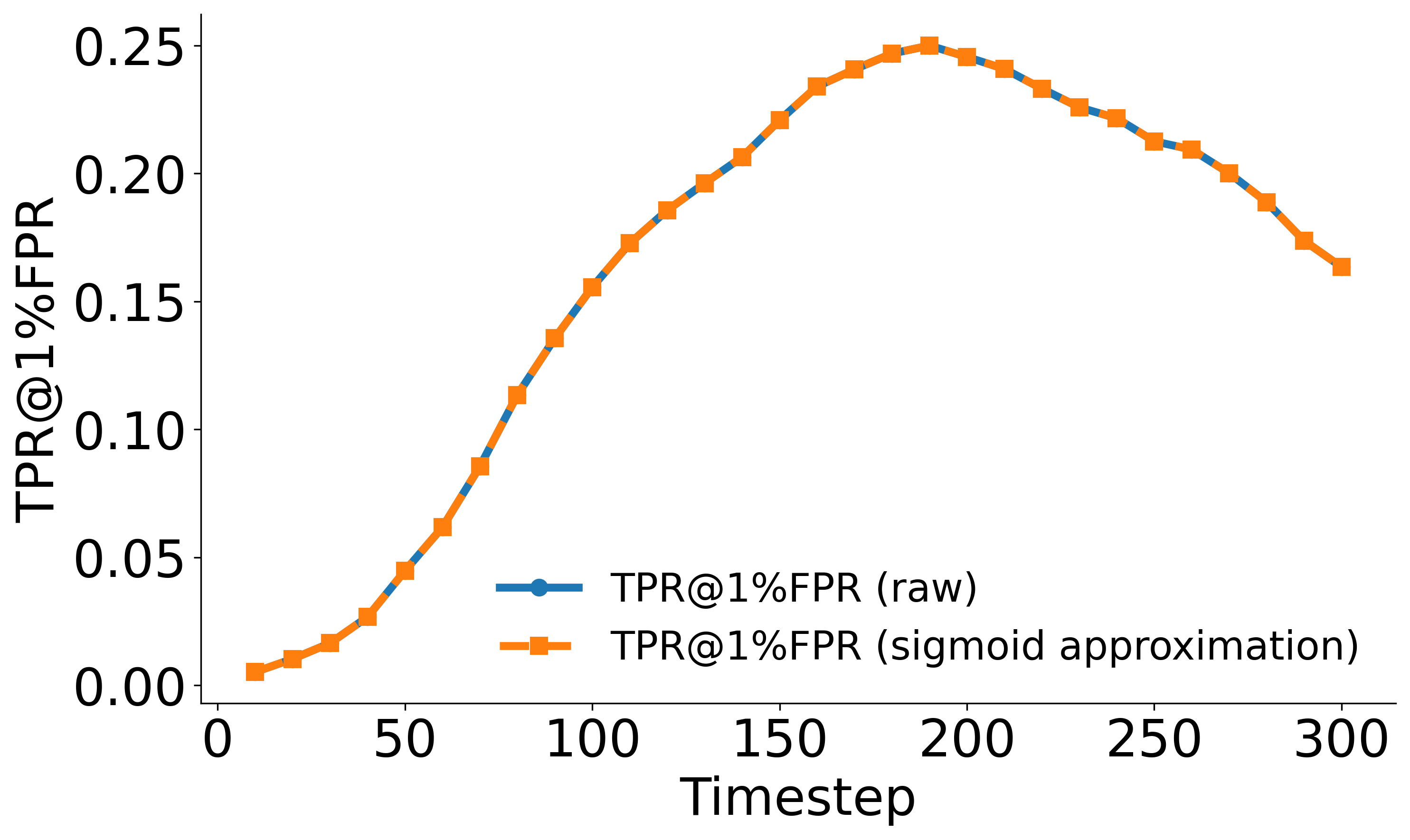}
        \caption{Comparison of TPR@1\%FPR under the original membership inference classifier and our approximation.}
        \label{fig:tpr_results}
    \end{subfigure}
    \caption{The comparison of MIA obtained under $\mathbf{1}[R_{t,p}(\boldsymbol{x}_{0})<\tau]$ and $\text{Sigmoid}(-(R_{t,p}(\boldsymbol{x}_{0}) - \tau) / \phi)$ under different timestep $t$, where $\tau = 12.5$, $p = 0.9$.}
    \label{fig:sigmoid_comparison}
    \vspace{-6mm}
\end{figure}

Owing to the discussion, we present the core idea of our method: a constructed membership inference classifier can be used to facilitate the conditional generation. Naturally, we refer to the membership inference classifier in \eqref{eq:attack_classify}. However, we observe that the classifier in \eqref{eq:bayes_rule} should be differentiable, whereas the classifier in \eqref{eq:attack_classify} is an indicator function. Thus, we propose to surrogate it with 
\begin{equation}
\begin{aligned}
    \textbf{1}[R_{t, p}(\boldsymbol{x}_{0}) < \tau] & \approx \text{Sigmoid}\left( - \frac{R_{t, p}(\boldsymbol{x}_{0}) - \tau}{\phi}\right) \\
    &= p(\boldsymbol{x}_{0}\in \boldsymbol{S}),
\end{aligned}
\label{eq:surrogate}
\end{equation}
where $\phi$ is a small constant. Figure \ref{fig:sigmoid_comparison} \footnote{TPR@1$\%$FPR means the true positive rate of membership inference under a false positive rate lower than 1$\%$, which is a standard criterion for verifying the effectiveness of MIA \cite{carlini2022membership}.} shows a comparison between the two membership inference attackers. The sigmoid provides a differentiable surrogate for the discrete membership decision. Because it is monotonic with respect to the membership score (with the sign chosen to match the decision rule), the corresponding ranking-based AUC is preserved. This agreement does not, by itself, validate probability calibration or the accuracy of the guidance gradients; the practical effect of the surrogate is assessed through extraction experiments. 

Furthermore, since
\begin{equation}
\begin{aligned}
& \nabla_{\boldsymbol{x}_{t}}\log{p(\boldsymbol{x}_{0}\in \boldsymbol{S} \mid \boldsymbol{x}_{t})} \\
& = \nabla_{\boldsymbol{x}_{t}}\log{\int p(\boldsymbol{x}_{0}\in \boldsymbol{S}, \boldsymbol{x}_{0} \mid \boldsymbol{x}_{t})d\boldsymbol{x}_{0}} \\
& = \nabla_{\boldsymbol{x}_{t}}\log{\int p(\boldsymbol{x}_{0}\in \boldsymbol{S})p(\boldsymbol{x}_{0} \mid \boldsymbol{x}_{t})d\boldsymbol{x}_{0}} \\
    & = \nabla_{\boldsymbol{x}_{t}}\log{\mathbb{E}_{\boldsymbol{x}_{0} \mid \boldsymbol{x}_{t}}[p(\boldsymbol{x}_{0} \in \boldsymbol{S})]} \\
    & \overset{\eqref{eq:x0_predict}}{\approx} \nabla_{\boldsymbol{x}_{t}}\log{p(\hat{\boldsymbol{x}}_{0}(\boldsymbol{x}_{t})\in\boldsymbol{S})} \\
    & \overset{\eqref{eq:surrogate}}{=} \nabla_{\boldsymbol{x}_{t}}\log\text{Sigmoid}\left(-\frac{R_{t,p}(\hat{\boldsymbol{x}}_{0}(\boldsymbol{x}_{t})) - \tau}{\phi}\right) \\
    & = f_{\phi}(\hat{\boldsymbol{x}}_{0}(\boldsymbol{x}_{t})),
\end{aligned}
\label{eq:guidance_signal}
\end{equation} 
the target guidance signal $\nabla \log p(\boldsymbol{x}_{0} \in \boldsymbol{S} \mid \boldsymbol{x}_{t})$ can be approximated by $f_{\phi}(\hat{\boldsymbol{x}}_{0}(\boldsymbol{x}_{t}))$ where 
\begin{equation}\label{eq:x0_predict}
    \hat{\boldsymbol{x}}_{0}(\boldsymbol{x}_{t}) = \frac{1}{\sqrt{\bar{\alpha}_{t}}}\left(\boldsymbol{x}_{t} - \sqrt{1 - \bar{\alpha_{t}}}\boldsymbol{\epsilon}_{\boldsymbol{\theta}}(\boldsymbol{x}_{t}, t)\right)
\end{equation}
is a one-step point estimate of the clean image given $\boldsymbol{x}_{t}$, rather than a sample drawn from the full posterior $p(\boldsymbol{x}_{0}\mid\boldsymbol{x}_{t})$. In particular, evaluating a nonlinear membership surrogate at this estimate is a tractable approximation to the conditional expectation above, not an exact probabilistic identity. Obtaining a more complete estimate would require repeated multistep denoising, increasing computational cost. Empirically, the one-step approximation yields more unique extractions and fewer duplicates than the evaluated multistep alternatives (Section \ref{sec:ablation}).

Thereafter, by plugging \eqref{eq:guidance_signal} into \eqref{eq:guided_noise}, we obtain the conditional noise prediction term 
\begin{equation}
    \hat{\boldsymbol{\epsilon}}_{\boldsymbol{\theta}}(\boldsymbol{x}_{t}, t) \approx  \boldsymbol{\epsilon}_{\boldsymbol{\theta}}(\boldsymbol{x}_{t}, t) - s\sqrt{1 - \bar{\alpha}_{t}}f_{\phi}(\hat{\boldsymbol{x}}_{0}(\boldsymbol{x}_{t})),
    \label{eq:conditional_noise}
\end{equation}
and use it to replace $\boldsymbol{\epsilon}_{\boldsymbol{\theta}}(\boldsymbol{x}_{t}, t)$ in DDIM \eqref{eq:ddim_denoising} to obtain samples from the training set $\boldsymbol{S}$, so that conduct the data extraction attack. 

Notably, similar to \cite{ho2022classifier}, we use a constant $s$ in \eqref{eq:conditional_noise} to rescale the magnitude of the guidance signal. Here, due to $\partial{\text{Sigmoid}(x)}/\partial{x} = \text{Sigmoid}(x)(1 - \text{Sigmoid}(x))\approx 0$, the chain rule results in $f_{\phi}(\boldsymbol{x}_{t})$ being a small scale. Thus, we should use a large $s$ in practice to strengthen the guidance signals. 

\subsection{Adversarial Guidance}
\label{sec:adversarial_guidance}

\begin{figure}[t]
    \centering

    \begin{subfigure}[b]{\columnwidth}
        \centering
        \includegraphics[
            width=\linewidth,
            trim=25 0 25 0,
            clip
        ]{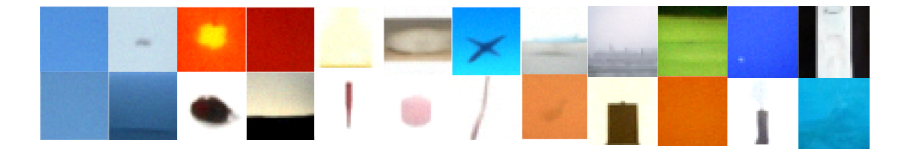}
        \caption{Generated images without adversarial guidance augmentation}
        \label{fig:without_augmentation}
    \end{subfigure}
    \begin{subfigure}[b]{\columnwidth}
        \centering
        \includegraphics[
            width=\linewidth,
            trim=25 0 25 0,
            clip
        ]{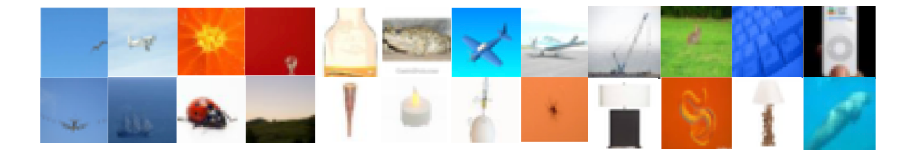}
        \caption{Generated images with adversarial guidance augmentation}
        \label{fig:with_augmentation}
    \end{subfigure}

    \caption{Comparison of images generated without and with adversarial
    guidance augmentation when constructing the guidance signal.}
    \label{fig:augmentation_comparison}
    \vspace{-6mm}
\end{figure}

As explored in \cite{goodfellow2014explaining,liu2024referee,papernot2016transferability,dinh2023rethinking}, extracting data from the training set may lead to the generation of ``adversarial samples'', which spuriously follow the guidance signal. This phenomenon severely breaks the effectiveness of data extraction attack on classification models as shown in \cite{yu2020cloudleak, he2021model, zhou2023boosting}.

Unfortunately, we find that adversarial samples also exist in our DEA on diffusion model. As shown in Figure \ref{fig:without_augmentation}, the conditional generation process extracts data that deceive the constructed membership inference classifier, while their poor details indicate that they are not from the training set. To address this, we propose constructing the guidance signal using adversarial gradient, similar to \cite{shen2024understanding}. Specifically, we replace $f_{\phi}(\hat{\boldsymbol{x}}_{0}(\boldsymbol{x}_{t}))$ with 
\begin{equation}
    f_{\phi}(\mathcal{T}(\hat{\boldsymbol{x}}_{0}(\boldsymbol{x}_{t}))) = \frac{1}{4}\sum_{i=1}^{4}f_{\phi}(\mathcal{T}_{i}(\hat{\boldsymbol{x}}_{0}(\boldsymbol{x}_{t}))),
    \label{eq:enhance_gradient}
\end{equation}
where $\{\mathcal{T}_{1}, \mathcal{T}_{2}, \mathcal{T}_{3}, \mathcal{T}_{4}\}$ is a set of random data augmentations, corresponding to color adjustments, resizing, translation, and cutout. By doing so, the constructed adversarial guidance signal ensures that the generated $\hat{\boldsymbol{x}}_{0}(\boldsymbol{x}_{t})$ satisfies the membership feature across multiple views, thereby mitigating the potential for generating adversarial samples. 

As visualized in Figure \ref{fig:augmentation_comparison}, with the adversarial guidance imposed, our DEA on diffusion generates samples with better details, mitigating the occurrence of adversarial samples.  

\subsection{Acceleration of Data Extraction Attack}
\label{sec:acceleration}
Although our proposed method improves the success rate of DEA on diffusion model, it also introduces additional computational overhead, i.e., the computation in the guidance signal $f_{\phi}(\mathcal{T}(\hat{\boldsymbol{x}}_{0}(\boldsymbol{x}_{t})))$. The computational process queries the PIA classifier multiple times during each sampling step, making the single data extraction process relatively time-consuming. Furthermore, excessive guidance during generation can degrade the quality of the generated image. To improve both efficiency and quality, we propose adding the guidance signal in a subset $\cG$ of the denoising steps, rather than at each step.  

Moreover, as shown in \cite{yi2024towards,yi2023generalization}, the denoising process will complete the details in the final stage of denoising, where the diversity of generations are appeared. This means that even starting from the same intermediate $\boldsymbol{x}_{t}$, the subsequent sampling can yield completely different results. Inspired by this observation, for a given initial $\boldsymbol{x}_{T}$, we propose to use DDIM \eqref{eq:ddim_denoising} on an initialization $\boldsymbol{x}_{T}$,  during denoising steps $[t_{0}, T]$ for some constant $0 \leq t_{0}\leq T$. Then, we repeat the obtained $\boldsymbol{x}_{t_{0}}$ for $m$ times to get $\{\boldsymbol{x}_{t_{0}}^{1}, \cdots, \boldsymbol{x}_{t_{0}}^{m}\}$, and parallel denoising these $\{\boldsymbol{x}_{t_{0}}^{i}\}_{i=1}^{m}$ by DDIM with some $\sigma > 0$ but under different random seed as shown in Figure \ref{fig:MGE_algo}. By utilizing the strong parallel computation ability of GPU, our parallel branch denoising strategy accelerates our data extraction attack.\footnote{Ideally, the parallel denoising strategy saves nearly $\frac{T - t_{0}}{T}$ computational cost.}  

\subsection{Membership Guided Extraction (MGE)}

\begin{algorithm}[h]
\footnotesize
\caption{Membership Guided Extraction (MGE)}
\label{alg:MGE_framework}
\begin{algorithmic}[1]
\Require 
$\boldsymbol{x}_T\sim \mathcal{N}(0, \boldsymbol{I})$, denoising steps $T$, branch number $m$, guidance steps $\mathcal{G}$, splitting time step $t_{0}$, noise scale $\sigma_{0}$, $\sigma_{1}$, 
\For{$t = T, \ldots, 1$}
    \If{$t\notin \mathcal{G}$}
    \State $\boldsymbol{\epsilon}_{t}^{*} = \boldsymbol{\epsilon}_{\boldsymbol{\theta}}(\boldsymbol{x}_{t}, t)$
    \Else
    \State $\boldsymbol{\epsilon}_{t}^{*} = \hat{\boldsymbol{\epsilon}}_{\boldsymbol{\theta}}(\boldsymbol{x}_{t}, t) = \boldsymbol{\epsilon}_{\boldsymbol{\theta}}(\boldsymbol{x}_{t}, t) - s\sqrt{1 - \bar{\alpha}_{t}}f_{\phi}(\mathcal{T}(\hat{\boldsymbol{x}}_{0}(\boldsymbol{x}_{t})))$
    \EndIf
    \If{$t > t_{0}$}
        \State $\boldsymbol{x}_{t - 1} = \mathrm{DDIM}(t, \boldsymbol{x}_{t}, \boldsymbol{\epsilon}_{t}^{*}, \sigma_{0})$
    \Else 
    \State Initialize $\boldsymbol{x}_{t_{0}}^{1} = \cdots = \boldsymbol{x}_{t_{0}}^{m} = \boldsymbol{x}_{t_{0}}$ when $t = t_{0}$
    \State Parallel Denoising for $i=1,\cdots, m$ under different seed
    \State $\boldsymbol{x}_{t - 1} = \mathrm{DDIM}(t, \boldsymbol{x}_{t}, \boldsymbol{\epsilon}_{t}^{*}, \sigma_{1})$
    \EndIf
\EndFor
\State \Return $\{\boldsymbol{x}_0^{1}, \ldots, \boldsymbol{x}_0^{m}\}$
\end{algorithmic}
\end{algorithm}

In summary, the three aforementioned techniques constitute our proposed DEA on diffusion i.e., Member Guided Extraction (MGE). In our MGE, the data extraction attack is transformed into a conditional generation process, and we construct the guidance signal by approximating a membership inference classifier. Moreover, we propose adversarial guidance and acceleration to improve the accuracy and the efficiency of our MGE. The overall process of MGE is shown in Algorithm \ref{alg:MGE_framework}. Furthermore, we present some extraction results in Figure \ref{fig:extracted}.

\section{Experiments}
\label{sec:experiments}

In this section, we evaluate the performance of our MGE on multiple datasets under different settings. The experimental results demonstrate the effectiveness of our method. 

\begin{table*}[h]
    \centering
    \resizebox{\textwidth}{!}{
    \begin{tabular}{c|ccc|ccc|ccc}
        \toprule
        \multicolumn{10}{c}{ADM} \\
        \midrule
        & \multicolumn{3}{c|}{\texttt{CIFAR10}} & \multicolumn{3}{c|}{\texttt{CIFAR100}} & \multicolumn{3}{c}{\texttt{Tiny-IN}} \\
        \midrule
        Method & Quantity $\uparrow$ & ASR $\uparrow$ & Efficiency $\uparrow$ & Quantity & ASR & Efficiency & Quantity & ASR & Efficiency \\
        \midrule
        MGE & $\boldsymbol{141}$ / $2^{17}$ & $\boldsymbol{0.107\%}$ & $\boldsymbol{0.78}$ & $\boldsymbol{358}$ / $2^{17}$ & $\boldsymbol{0.273\%}$ & $\boldsymbol{1.94}$ & $\boldsymbol{163}$ / $2^{17}$ & $\boldsymbol{0.124\%}$ & $\boldsymbol{0.88}$ \\
        DDPM \cite{carlini2023extracting} & 55 / $2^{17}$ & 0.042\% & 0.08 & 129 / $2^{17}$ & 0.098\% & 0.18 & 41 / $2^{17}$ & 0.031\% & 0.06 \\
        DDIM \cite{carlini2023extracting} & 26 / $2^{17}$ & 0.020\% & 0.37 & 52 / $2^{17}$ & 0.040\% & 0.57 & 24 / $2^{17}$ & 0.018\% & 0.25 \\
        \midrule
        \multicolumn{10}{c}{EDM} \\
        \midrule
        & \multicolumn{3}{c|}{\texttt{CIFAR10}} & \multicolumn{3}{c|}{\texttt{AFHQv2}} 
        & \multicolumn{3}{c}{\texttt{FFHQ}} \\
        \midrule
        Method & Quantity & ASR & Efficient & Quantity & ASR & Efficient & Quantity & ASR & Efficient \\
        \midrule
        MGE & $\boldsymbol{46}/2^{17}$ & $\boldsymbol{0.035\%}$ & $\boldsymbol{0.51}$ 
            & $\boldsymbol{123}/2^{17}$ & $\boldsymbol{0.094\%}$ & $\boldsymbol{0.41}$ 
            & $\boldsymbol{23}/2^{17}$ & $\boldsymbol{0.018\%}$ & $\boldsymbol{0.10}$ \\
        DDPM \cite{carlini2023extracting} & $22 / 2^{17}$ & 0.017\% & 0.04
            & $17 / 2^{17}$ & 0.013\% & 0.04 
            & $4 / 2^{17}$ & 0.003\% &  0.01 \\
        DDIM \cite{carlini2023extracting} & $17 / 2^{17}$ & 0.013\% & 0.23 
            & $27 / 2^{17}$ & 0.021\% & 0.22 
            & $2 / 2^{17}$  & 0.002\% & 0.02 \\
        \midrule
        \multicolumn{10}{c}{Stable Diffusion v1.5} \\ 
        \midrule
        Method & \multicolumn{3}{c|}{Quantity $(500 \times 500)$} & \multicolumn{3}{c|}{ASR} & \multicolumn{3}{c}{Efficiency} \\
        \midrule
        MGE & \multicolumn{3}{c|}{$\boldsymbol{27}$} & \multicolumn{3}{c|}{$\boldsymbol{0.0108\%}$} & \multicolumn{3}{c}{$\boldsymbol{0.075}$} \\
        DDPM \cite{carlini2023extracting} & \multicolumn{3}{c|}{7} & \multicolumn{3}{c|}{0.0028\%} & \multicolumn{3}{c}{0.001} \\
        DDIM \cite{carlini2023extracting} & \multicolumn{3}{c|}{5} & \multicolumn{3}{c|}{0.0020\%} & \multicolumn{3}{c}{0.033} \\
        \bottomrule
    \end{tabular}
    }
    \caption{Performance comparison of different data extraction methods under different datasets and diffusion models. The low ASR of data extraction attack of stable diffusion v1.5 is due to the large generation diversity of this Text2Image model \cite{ramesh2022hierarchical}}
    \label{tb:main_experiment}
    \vspace{-6mm}
\end{table*}

\subsection{Set Up}
\label{sec:set up}

\paragraph{Evaluation Metrics.} For the data extraction attack, \textbf{Quantity} is the number of distinct extracted training examples after deduplication and verification. We report \textbf{ASR} (Attack Success Rate) as the number of unique extractions divided by the total number of generated candidates, rather than the frequency of repeated reproductions. \textbf{Efficiency} is the number of unique extractions per minute, counting the full generation process including guidance computation. The experiments use batch size 4. All these results are conducted on a workstation with a single NVIDIA-A800 GPU.

\paragraph{Benchmarks and Baselines.} We compare our MGE to the baseline method ``generate-and-filter'' in \cite{carlini2023extracting} on several image generation benchmark datasets, including \texttt{CIFAR10}, \texttt{CIFAR100} \cite{krizhevsky2009learning}, \texttt{Tiny-ImageNet} (\texttt{Tiny-IN}) \cite{deng2009imagenet}, \texttt{FFHQ} \cite{karras2019style}, \texttt{AFHQv2} \cite{choi2020stargan}, and \texttt{LAION-2B} \cite{schuhmann2022laion}.

\paragraph{Hyperparameters.} In our MGE (Algorithm \ref{alg:MGE_framework}), we use a 100-step DDIM for conditional generation, with a guidance scale $s = 70.0$, branch number $m = 4$. The guidance subset is $\mathcal{G}=\{4t: 5\leq t \leq 16\}$. We set the splitting point $t_{0} = 50$, using DDIM ($\sigma_{0} = 1.0$) in the first denoising stage $[t_{0}, T]$, and DDIM ($\sigma_{1} = 1.0$) in the second denoising stage $[0, t_{0}]$. We evaluate our method on diffusion models  ADM \cite{nichol2021improved}, EDM \cite{karras2022elucidating}, and stable-diffusion v1.5 \cite{ramesh2022hierarchical} (trained on \texttt{LAION-5B}). We refer readers to Appendix \ref{app:setup} for more details.

\paragraph{Detection of Extracted Data.} For \texttt{CIFAR10}, \texttt{CIFAR100}, \texttt{Tiny-IN}, \texttt{FFHQ} and \texttt{AFHQv2}, they are publicly available datasets in a relative small size. Thus, following \cite{carlini2023extracting}, we detect whether a generated data $\hat{\boldsymbol{x}}$ is in the training set by evaluating its distance with all $\boldsymbol{x}$ from training set as  
\begin{equation}
    \ell\left(\hat{\boldsymbol{x}}, \boldsymbol{x} ; N_{\hat{\boldsymbol{x}}}\right) = \frac{\ell_{2}(\hat{\boldsymbol{x}}, \boldsymbol{x})}{\alpha \cdot \frac{1}{|N_{\hat{\boldsymbol{x}}}|} \sum_{y \in N_{\hat{\boldsymbol{x}}}} \ell_{2}(\hat{\boldsymbol{x}}, \boldsymbol{y})},
    \label{eq:improved_l2}
\end{equation}
where $N_{\hat{\boldsymbol{x}}}$ is the set containing the $|N_{\hat{\boldsymbol{x}}}|$ closest samples from the training dataset to the example $\hat{\boldsymbol{x}}$. A smaller $\ell(\hat{\boldsymbol{x}}, \bx, N_{\hat{\bx}})$ indicates that the training sample $\boldsymbol{x}$ is closer to the generated $\hat{\boldsymbol{x}}$ than its neighboring samples in the training set, implying that $\hat{\boldsymbol{x}}$ is more likely to belong to the training set. The detection method is shown to have very high accuracy in \cite{carlini2023extracting}. Following \cite{carlini2023extracting}, we set $\alpha = 0.5$, $|N_{\hat{\boldsymbol{x}}}| = 50$, and recognize $\hat{\bx}$ is from training set when $\ell\left(\hat{\boldsymbol{x}}, \boldsymbol{x} ; N_{\hat{\boldsymbol{x}}}\right) < 0.9$. After screening candidates with the normalized distance criterion, we manually verify potential extractions and remove duplicates before counting unique extracted instances. The same dataset-specific threshold and manual verification procedure are used for MGE and the corresponding baselines. For Text2Image model Stable Diffusion v1.5, since its complete training set \texttt{LAION-5B} is unavailable, we extract data from \texttt{LAION-2B} (a subset of \texttt{LAION-5B}) and follow the detection method (with verified high accuracy) in \cite{carlini2023extracting}. More details of the detection method refers to Appendix \ref{app:setup}.

\subsection{Main Results} 

We evaluate the Quantity, ASR, and Efficiency for data extraction attacks in Table \ref{tb:main_experiment}. Our MGE method is compared with baseline method in \cite{carlini2023extracting}, using DDIM ($\sigma = 0$) and DDPM \cite{ho2020denoising} for sampling. DDIM offers higher generation efficiency, while DDPM achieves greater generation diversity. We generate $2^{17}$ images and compute the evaluation metrics defined in \eqref{eq:improved_l2}. From the empirical results in Table \ref{tb:main_experiment}, we have the following observations. 

\begin{enumerate}
    \item From the comparison between baseline methods, DDPM tends to generate more training data compared to DDIM, owing to its higher generation diversity caused by injected noise. However, DDPM is less efficient than DDIM because of its longer denoising process. In contrast, by our parallel branch denoising, our MGE inherits the advantages of both DDPM and DDIM, achieving a good balance between efficiency and diversity. 
    \item The proposed MGE significantly outperforms the baselines in Quantity, ASR, and Efficiency, demonstrating that converting the data extraction attack into a conditional generation process brings substantial benefits. 
\end{enumerate}

A separate comparison with SIDE is reported in
Appendix \ref{app:side_comparison}.
Additionally, a fully matched unguided control
is included in Table \ref{tb:scale_impact},
where setting $s=0$ disables membership guidance
while keeping the sampling and branching
configurations unchanged.

\subsection{Ablation Study}\label{sec:ablation}

Next, we investigate the effectiveness of different components in our MGE. To this end, we conduct a series of ablation studies on the \texttt{CIFAR10}. Additional analyses on hyperparameter sensitivity, Stable Diffusion component ablations, and random-seed variability are provided in the Appendix (Sections \ref{app:sd_ablation} and \ref{app:seed_stability}).

\paragraph{Effect of different MIA classifier.} Since our MGE framework formulates data extraction as a conditional generation process guided by a membership inference attacker, it is intuitive that the capability of the underlying MIA classifier directly influences the extraction performance. To systematically investigate this dependence, we integrate MGE with different MIA classifiers, including Naive Attack (NA) \cite{duan2023diffusion}, SecMI \cite{duan2024membership}, and Proximal Initialization Attack (PIA) \cite{kong2023efficient} used in our paper. The extraction results on CIFAR10 are presented in Table \ref{tb:ablation_mia}.

Among the evaluated MIA methods, stronger membership discrimination is associated with higher extraction success rates: NA, SecMI, and PIA achieve ASRs of 0.025\%, 0.053\%, and 0.107\%, respectively, consistent with their AUC and TPR@1\%FPR rankings. Efficiency, however, does not follow the same monotonic trend, since each MIA incurs a different guidance-computation cost. These observations show that MGE can integrate different MIA methods, while extraction quality and computational efficiency depend on different aspects of the chosen guidance signal.

\begin{table*}[h]
    \centering
    \begin{tabular*}{\textwidth}{@{\extracolsep{\fill}}c|ccccc@{}}
        \toprule
        MIA Classifier & Quantity & ASR & Efficiency & AUC & TPR@1\%FPR \\
        \midrule
        MGE-NA & 33 / $2^{17}$ & 0.025\% & 0.22 & 84.7 & 6.85 \\
        MGE-SecMI & 70 / $2^{17}$ & 0.053\% & 0.03 & 88.1 & 9.11 \\
        MGE-PIA & 141 / $2^{17}$ & 0.107\% & 0.78 & 91.9 & 24.7\\
        \bottomrule
    \end{tabular*}
    \caption{MGE under varied MIAs on CIFAR-10 with AUC and TPR@1\%FPR of MIAs reported.}
    \label{tb:ablation_mia}
\end{table*}

\subsubsection{Ablation Study on Adversarial Guidance} 

\paragraph{Effect of Random Augmentation.} As mentioned in Section \ref{sec:adversarial_guidance}, the guidance signal is constructed through random data augmentation to obviate adversarial sample \cite{goodfellow2014explaining,arrieta2020explainable}. Next, we vary the augmentation strength to analyze its impact on MGE. The results are shown in Table \ref{tb:augmentation_impact}. 

As can be seen, increasing the number of data augmentation operators in our adversarial guidance improves the accuracy of the guidance signal but introduces additional computational overhead due to multiple forward passes. This accuracy v.s. efficiency trade-off reflects that: an appropriate number of augmentation operations should be empirically determined in practice to balance them.

\begin{table}[H]
    \centering
    \resizebox{\columnwidth}{!}{
    \begin{tabular}{c|ccc}
        \toprule
        Augmentation Strength & Quantity & ASR & Efficiency \\
        \midrule
        without augmentation & 18 / $2^{17}$ & 0.014\% & 0.2 \\
        1 & 58 / $2^{17}$ & 0.044\% & $\boldsymbol{0.64}$ \\
        2 & 57 / $2^{17}$ & 0.043\% & 0.48 \\
        4 & $\boldsymbol{141}$ / $2^{17}$ & $\boldsymbol{0.107\%}$ & 0.78 \\
        \bottomrule
    \end{tabular}
    }
    \caption{Performance of MGE under varied composition of augmentation operations on \texttt{CIFAR10}.}
    \label{tb:augmentation_impact}
\end{table}

\subsubsection{Ablation Study on Acceleration} In our MGE, we use three tricks, i.e., guidance in a subset (Section \ref{sec:acceleration}), parallel branch denoising (Section \ref{sec:acceleration}), and one-step prediction \eqref{eq:x0_predict}. Next, we explore their effects to our MGE. 

\paragraph{Effect of Parallel Branch Denoising.} To further accelerate MGE, we propose a parallel branch denoising technique. We explore its effectiveness by varying the number of branches, as summarized in Table \ref{tb:branch_impact}, increasing the number of branches improves  \footnote{Theoretically, the decreased computational cost is upper bounded by $\frac{T - t_{0}}{T}$.} the efficiency of MGE but reduces the diversity of generated samples when branch number is large e.g., $\geq 8$. Excessive parallelism reduces stochastic variation in denoising, making generation more deterministic. Thus, a moderate branch count better balances efficiency and diversity. 

\begin{table}[H]
    \centering
    \begin{tabular}{@{\extracolsep{\fill}}ccccc@{}}
        \toprule 
        Branch Nums & 1 & 2 & 4 & 8 \\
        \midrule
        Quantity & 73 / $2^{17}$ & 60 / $2^{17}$ & $\mathbf{141}$ / $2^{17}$ & 65 / $2^{17}$ \\
        ASR & 0.056\% & 0.046\% & $\mathbf{0.107\%}$ & 0.050\% \\
        Efficiency & 0.35 & 0.36 & $\mathbf{0.78}$ & 0.50 \\
        \bottomrule
    \end{tabular}
    \caption{Performance of MGE with varied branch number under \texttt{CIFAR10}.}
    \label{tb:branch_impact}
\end{table}

\paragraph{Effect of Guidance Approximation.} 
\label{guidance_approximation}
During MGE, we use the one-step clean-image estimate in \eqref{eq:x0_predict} to approximate the guidance objective, rather than sampling from the full posterior $p(\bx_{0}\mid\bx_{t})$.   
Next, we compare it against several non-approximate DDIM baselines with varying precision levels, i.e., denoising $x_{t}$ with 1/5 total steps, 1/2 total steps, and full steps. As shown in Table \ref{tb:ablation_approximate}, we observe a counter-intuitive phenomenon: increasing guidance precision leads to severe ``over-memorization'' and mode collapse. Specifically, the evaluated multistep alternatives yield only 15--18 unique extractions, compared with 141 for one-step prediction, and exhibit substantially more duplicates (over 13,800 versus 1,094). These observations are consistent with sampling trajectories concentrating on a smaller set of memorized examples; they do not establish that approximation error itself causes higher diversity. Compared to standard DDIM denoising, our one-step prediction trick improves the efficiency more than 40$\times$, which indicates it is not merely a speed-up technique, but also a superior extraction strategy. 

\begin{table}[H]
    \centering
    \small
    \setlength{\tabcolsep}{3.5pt}
    \begin{tabular*}{\columnwidth}{
        @{\extracolsep{\fill}}ccccc@{}
    }
        \toprule
        \multicolumn{1}{c}{
            \begin{tabular}[c]{@{}c@{}}
                Approximation\\
                Strategy
            \end{tabular}
        }
        & Quantity & ASR & Efficiency & Duplicates \\
        \midrule
        \textbf{Ours}
        & $141/2^{17}$ & 0.107\% & 0.783 & 1094 \\
        1/5 total steps
        & $15/2^{17}$ & 0.011\% & 0.020 & 13859 \\
        1/2 total steps
        & $16/2^{17}$ & 0.012\% & 0.011 & 13844 \\
        Full steps
        & $18/2^{17}$ & 0.014\% & 0.007 & 13885 \\
        \bottomrule
    \end{tabular*}
    \caption{Performance of MGE under varied approximation strategies.}
    \label{tb:ablation_approximate}
\end{table}

Table \ref{tb:ablation_approximate} indicates that more extensive multistep estimation does not necessarily improve unique training-data extraction. One possible explanation is that stochasticity and approximation affect the diversity of sampling trajectories, although these experiments do not isolate the causal role of approximation error. Additional component-level results on Stable Diffusion v1.5 are given in Appendix \ref{app:sd_ablation}.

\section{Conclusion}
\label{sec:conclusion}
In this paper, we further investigate the data privacy vulnerabilities of diffusion models and propose a novel data extraction attack, termed \textbf{Membership Guidance Extraction} (\textbf{MGE}). MGE formulates the attack as a membership guided conditional generation process, with three key components: a membership inference classifier, adversarial guidance, and accelerated guidance estimation. 

Extensive experiments on multiple benchmarks demonstrate the effectiveness of our method. Compared with the evaluated generate-and-filter approaches \cite{carlini2023extracting}, MGE uses membership-derived conditional guidance to direct generation without training an additional auxiliary classifier. Additional comparisons with SIDE, matched unguided controls, zeroth-order gradient estimation, and Stable Diffusion component ablations are presented in the appendix. These findings highlight that, before the powerful generative models can be safely deployed across various domains, it is essential to develop privacy-preserving mechanisms grounded in solid theoretical foundations to defend against such active attacks. 

\small
\bibliography{aaai2027}


\clearpage
\appendix

\section{Additional Experimental Results and Ablation Studies}

\subsection{Ablation Study on Guidance Scale}
Our MGE is built upon the MIAs, next we explore the effect of guidance scale to our MGE. 
\paragraph{Effect of Guidance Scale.} As shown in \eqref{eq:guidance_signal}, the magnitude of the guidance scale is controlled by the hyperparameter $s$. In Table \ref{tb:scale_impact}, we vary $s$ to examine its impact. As observed, the strength of MGE first increases and then decreases as $s$ grows. Intuitively, a larger guidance scale steers the generation process closer to the training set, but may degrade the visual quality of the generated images. Therefore, selecting an appropriate guidance scale allows MGE to achieve a good balance between fidelity and quality in practice. 

\begin{table*}[h]
    \centering
    \resizebox{\textwidth}{!}{
        \begin{tabular}{cccccccccc}
        \toprule 
        Guidance Scale & 0.0 & 10.0 & 20.0 & 40.0 & 50.0 & 60.0 & 70.0 & 80.0 & 90.0 \\
        \midrule
        Quantity & 6 / $2^{17}$ & 14 / $2^{17}$ & 39 / $2^{17}$ & 72 / $2^{17}$ & 71 / $2^{17}$ & 70 / $2^{17}$ & $141$ / $2^{17}$ & 58 / $2^{17}$ & 56 / $2^{17}$ \\
        ASR  & 0.005\% & 0.011\% & 0.030\% & 0.055\% & 0.054\% & 0.053\% & $0.107\%$ & 0.044\% & 0.043\% \\
        Efficiency & 0.03 & 0.08 & 0.22 & 0.40 & 0.39 & 0.39 & $0.78$ & 0.32 & 0.31 \\
        \bottomrule
    \end{tabular}
    }
    \caption{Performance of MGE under varied guidance scale $s$ on \texttt{CIFAR10}.}
    \label{tb:scale_impact}
\end{table*}

\subsection{Ablation Study on Temperature $\phi$}
In Section 4.1, we approximate the non-smooth membership indicator $1[R_{t,p}(x_0) < \tau]$ using a differentiable surrogate, i.e., $\text{Sigmoid}(-(R_{t,p}(x_0)-\tau)/\phi)$, where $\phi$ controls the smoothness and the precision of this approximation. To evaluate its impact, we conduct an ablation study by varying the inverse temperature $1/\phi$ from $1.0$ to $15.0$ on the CIFAR10 dataset.

\begin{table}[H]
    \centering
    \small
    \setlength{\tabcolsep}{3.5pt}
    \renewcommand{\arraystretch}{1.05}

    \begin{tabular*}{\columnwidth}{
        @{\extracolsep{\fill}}cccccc@{}
    }
        \toprule
        $1/\phi$ & 1.0 & 2.0 & 5.0 & 10.0 & 15.0 \\
        \midrule
        Quantity
        & $11/2^{17}$
        & $11/2^{17}$
        & $26/2^{17}$
        & $54/2^{17}$
        & $141/2^{17}$ \\

        ASR (\%)
        & 0.010
        & 0.010
        & 0.020
        & 0.041
        & 0.107 \\

        Efficiency
        & 0.06
        & 0.06
        & 0.15
        & 0.31
        & 0.78 \\

        AUC
        & 91.7
        & 91.8
        & 91.9
        & 91.9
        & 91.9 \\

        TPR@1\%FPR
        & 24.6
        & 24.7
        & 24.7
        & 24.7
        & 24.7 \\
        \bottomrule
    \end{tabular*}
    \caption{MGE on the CIFAR-10 dataset under various $\phi$.}
    \label{tb:analysis_phi}
\end{table}

As summarized in Table \ref{tb:analysis_phi}, the effect of our MGE is increasing with $1/\phi$ (i.e., as $\phi \rightarrow 0$). Since a smaller $\phi$ makes the continuous surrogate more closely approximate the discrete membership indicator, it further confirms that more confident membership predictions in MGE lead to improved data extraction performance. Here, for the binary classification problem membership inference, the improvements of surrogate is not revealed by AUC, since the smaller $1/\phi$ change the confidence (decision is made by comparing it with 0.5) on classification, instead of its results.  

\subsection{Ablation Study on Guidance Phase Selection}
In MGE, the guidance signal is injected at denoising steps denoted by $\mathcal{G}$. We vary the selection of $\mathcal{G}$ to examine its influence on MGE performance, with results summarized in Table \ref{tb:denoising_steps_impact}. As shown, applying guidance in the middle denoising phase ($\mathcal{G}=\{4t: 5\leq t \leq 16\}$) achieves the best results, yielding the highest ASR and high efficiency. When applied only in the early or late stages, both extraction accuracy and efficiency drop sharply. 

We speculate this originates from the ``first shape then overall details'' diffusion process \cite{yi2024towards}. In this regime, early-stage guidance disrupts generation quality, while late-stage guidance has little influence. These results indicate that carefully choosing the guidance phase is critical for our MGE.

\begin{table}[H]
    \centering
    \resizebox{\columnwidth}{!}{
    \begin{tabular}{c|ccc}
        \toprule
        Guidance Phase & Quantity & ASR & Efficiency \\
        \midrule
        $\mathcal{G}=\{4t: 5\leq t \leq 16\}$ 
        & $\boldsymbol{141}$ / $2^{17}$ 
        & $\boldsymbol{0.107\%}$ 
        & $\boldsymbol{0.78}$ \\

        $\mathcal{G}=\{4t: 12\leq t \leq 25\}$ 
        & 42 / $2^{17}$ 
        & 0.032\% 
        & 0.42 \\

        $\mathcal{G}=\{4t: 5\leq t \leq 12\}$ 
        & 7 / $2^{17}$ 
        & 0.005\% 
        & 0.03 \\

        $\mathcal{G}=\{4t: 0\leq t \leq 25\}$ 
        & 47 / $2^{17}$ 
        & 0.036\% 
        & 0.20 \\
        \bottomrule
    \end{tabular}
    }
    \caption{Performance of MGE under varied guidance phases $\mathcal{G}$ across different diffusion timesteps $t$ on \texttt{CIFAR10}.}
    \label{tb:denoising_steps_impact}
\end{table}

\subsection{Ablation Study on Generation Number}

\begin{figure}[t]
    \centering
    \includegraphics[width=\columnwidth]{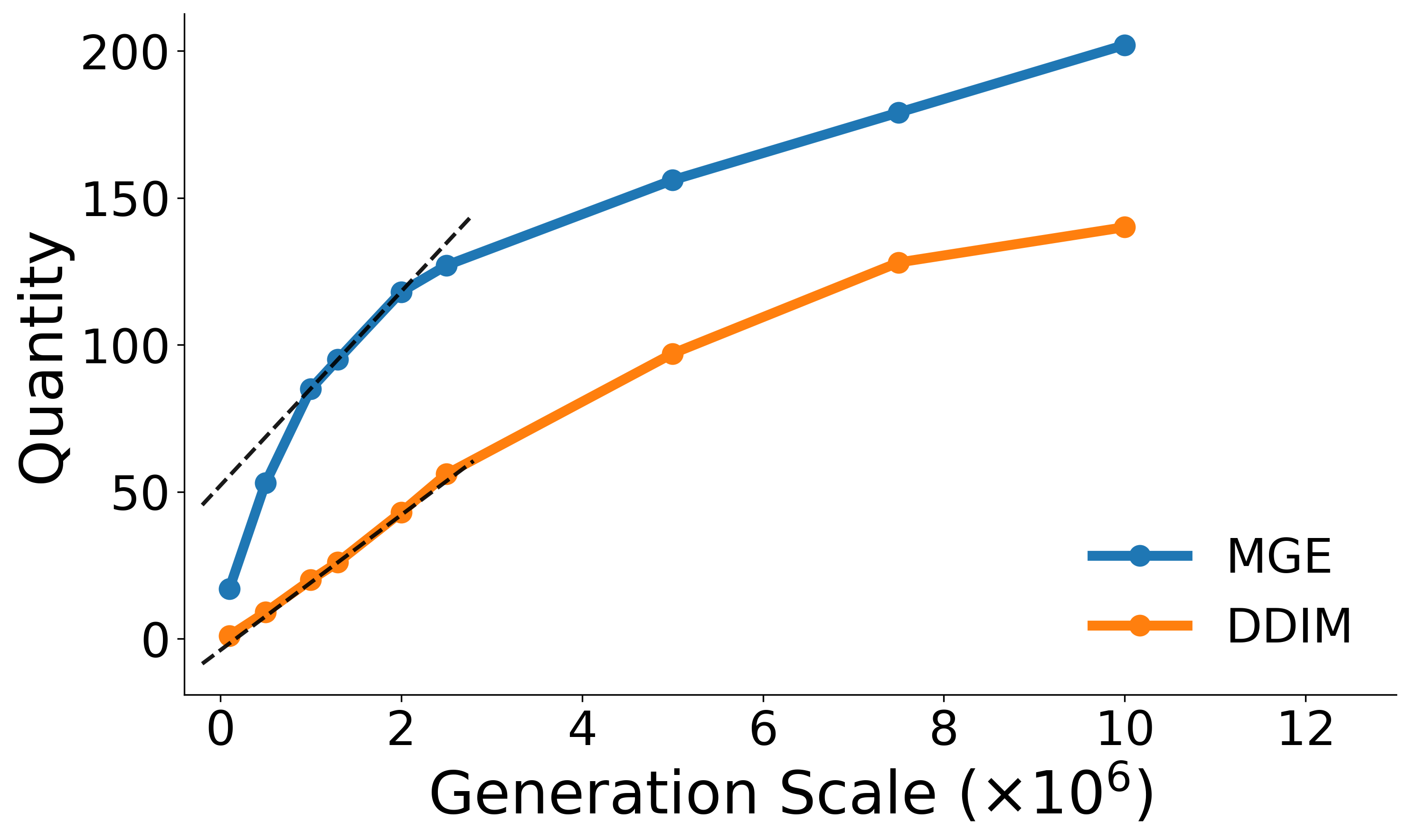}
    \caption{Comparison of expected and actual extraction quantities under different generation scales.}
    \label{fig:generation_scale}
\end{figure}

\begin{figure}[t]
    \centering
    \includegraphics[width=\columnwidth]{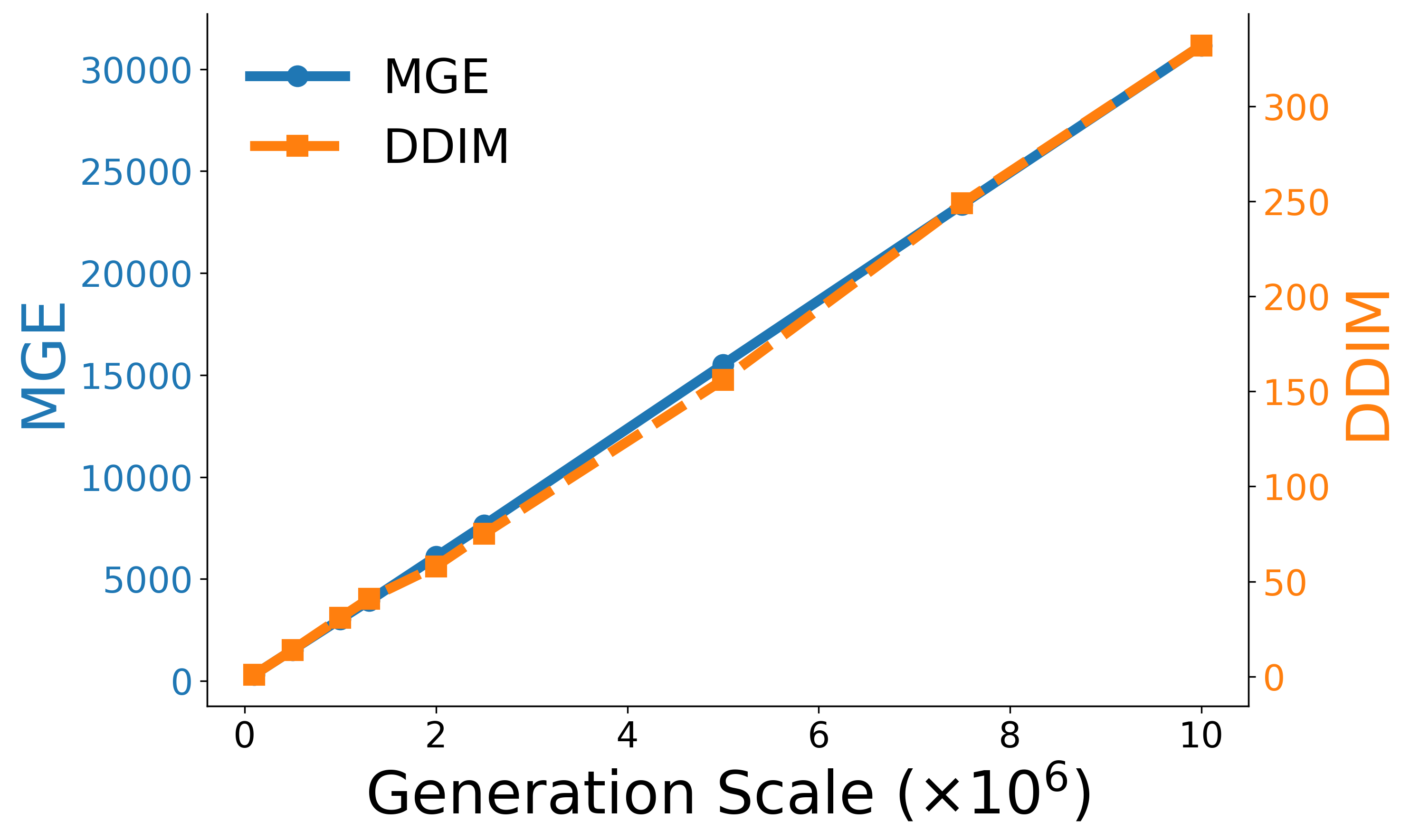}
    \caption{Variation in extracted images (non-deduplicated) across different generation scales.}
    \label{fig:repetitions_number}
\end{figure}

In Table \ref{tb:main_experiment}, each group produces $2^{17}$ candidates on the CIFAR10 dataset. As illustrated in Figures \ref{fig:generation_scale} and \ref{fig:repetitions_number}, the ASR drops sharply as the generation scale increases for both MGE and the baseline. This trend is primarily due to the diffusion model generating highly similar samples at larger scales, leading to significant output redundancy and mode collapse. This suggests that certain training samples are more easily ``memorized", causing the generation distribution to skew toward them. Furthermore, Figure \ref{fig:repetitions_number} indicates that while MGE extracts far more samples than the DDIM baseline, a substantial portion of these are duplicates.

As these memorized samples begin to dominate the output, additional generations yield diminishing returns. Reducing such over-memorization may further improve extraction efficiency, which we leave for future work.

\paragraph{Deduplicate Methods.} In order to explore ways to mitigate the aforementioned collapse, We introduce deduplication guidance based on nearest neighbor distances during the sampling process. 

In our deduplication algorithm, we maintain a queue that stores the images generated from previous sampling. Specifically, at each selected time step $t$, we approximate $\hat{\bx}_0$ from the current state $\bx_t$ according to \eqref{eq:x0_predict}, and then calculate the Euclidean distance $dist_i = \sqrt{\Vert a_i - b_i \Vert^2 / d}$ between $\hat{\bx}_0$ and each image in the queue. After that, we introduce $\tau_{dup}$ and use the sigmoid function to map the distance $dist_i$ to the repetition probability $p_i^{dup} = \text{Sigmoid}(-dist_i + \tau_{dup})$. We then define the de-duplication loss as
\begin{equation}
    \mathcal{L}_{dup} = -\log\bigl(1 - p_{i^*}^{dup}\bigr), \quad i^* = \arg\min_{i} \, dist_i,
\label{deduplication_loss}
\end{equation}
where $i^*$ denotes the index of the nearest neighbor in the queue. The gradient $\nabla_{\hat{\bx_0}} \mathcal{L}_{dup}$ is projected back into the noise prediction space and added to the original guidance terms. This method prevents the generated samples from collapsing into the neighborhoods of historical samples, thereby improving the diversity of the results without modifying the diffusion backbone.

As can be seen in Table \ref{tab:deduplication_comparison}, we find that while our proposed method for mitigating model collapse is effective to some extent, it significantly reduces generation efficiency. Therefore, further research is needed to explore this issue.

\begin{table}[H]
    \centering
    \small
    \setlength{\tabcolsep}{3.5pt}
    \begin{tabular}{ccccc}
        \toprule
        Setting & Quantity & ASR & Efficiency & Duplicates \\
        \midrule
        Non-deduplicated & $26/2^{14}$ & 0.159\% & 1.18 & 775 \\
        Deduplicated     & $28/2^{14}$ & 0.171\% & 0.20 & 548 \\
        \bottomrule
    \end{tabular}
    \caption{Performance comparison between non-deduplicated and deduplicated settings.}
    \label{tab:deduplication_comparison}
\end{table}

\subsection{Comparison with SIDE}
\label{app:side_comparison}
To compare MGE with another guided extraction approach, we reproduce SIDE under the evaluated CIFAR-10 setting. As shown in Table~\ref{tab:side_comparison}, MGE obtains more unique extractions and higher ASR and Efficiency. SIDE uses cluster-based surrogate conditions, whereas MGE directly constructs guidance from model-derived membership signals. These results provide an empirical comparison for this setting and should not be interpreted as a universal ranking across all attack settings.

\begin{table}[H]
    \centering
    \small
    \begin{tabular}{lccc}
        \toprule
        Method & Quantity & ASR & Efficiency \\
        \midrule
        SIDE & 96 & 0.073\% & 0.20 \\
        MGE & \textbf{141} & \textbf{0.107\%} & \textbf{0.78} \\
        \bottomrule
    \end{tabular}
    \caption{Comparison with reproduced SIDE results on CIFAR-10.}
    \label{tab:side_comparison}
\end{table}

\subsection{Zeroth-Order Guidance Estimation}
\label{app:zeroth_order}
The default MGE implementation obtains guidance gradients by backpropagation. We also test a zeroth-order alternative using two-sided finite differences to approximate the gradient of a queryable guidance objective $f$:
\begin{equation}
\nabla \hat{f}(\bx)=\frac{1}{K}\sum_{i=1}^{K}
\frac{f(\bx+\mu\bu_i)-f(\bx-\mu\bu_i)}{2\mu}\bu_i,
\quad \bu_i\sim\mathcal{N}(0,I),
\label{eq:zeroth_order}
\end{equation}
where $K$ is the number of directions and $\mu$ is the finite-difference perturbation scale. The results in Table~\ref{tab:zeroth_order} show that the zeroth-order variant retains 118 of the 141 unique extractions achieved with backpropagation (approximately 84\%), but at reduced efficiency due to additional objective evaluations. This experiment indicates that explicit backpropagation is not strictly required when the guidance objective can be queried at perturbed inputs; it does not establish an attack using only a standard image-generation API.

\begin{table}[H]
    \centering
    \small
    \begin{tabular}{lccc}
        \toprule
        Guidance gradient & Quantity & ASR & Efficiency \\
        \midrule
        Zeroth-order & 118 & 0.090\% & 0.44 \\
        Backpropagation & \textbf{141} & \textbf{0.107\%} & \textbf{0.78} \\
        \bottomrule
    \end{tabular}
    \caption{Gradient-estimation comparison on CIFAR-10.}
    \label{tab:zeroth_order}
\end{table}

\subsection{Stable Diffusion v1.5 Component Ablations}
\label{app:sd_ablation}
We conduct component-level evaluations on Stable Diffusion v1.5 to examine whether the benefits of one-step prediction and parallel branching extend beyond CIFAR-10. As shown in Table~\ref{tab:sd_ablation}, full MGE yields 27 unique extractions, with an ASR of 0.0108\% and an efficiency of 0.075. Using multistep prediction with one-fifth of the total denoising steps reduces the results to 5, 0.0020\%, and 0.003. Removing branching yields 14, 0.0056\%, and 0.032. These results support the usefulness of both components in this evaluated text-to-image setting.

\begin{table}[H]
    \centering
    \small
    \begin{tabular}{lccc}
        \toprule
        Setting & Quantity & ASR & Efficiency \\
        \midrule
        Full MGE & \textbf{27} & \textbf{0.0108\%} & \textbf{0.075} \\
        Multistep (1/5 steps) & 5 & 0.0020\% & 0.003 \\
        Without branching & 14 & 0.0056\% & 0.032 \\
        \bottomrule
    \end{tabular}
    \caption{Component ablations on Stable Diffusion v1.5.}
    \label{tab:sd_ablation}
\end{table}

\subsection{Stability Across Random Seeds}
\label{app:seed_stability}
We report the number of unique extractions from three CIFAR-100 runs with different random seeds in Table~\ref{tab:seeds}. The results are 358, 346, and 351, corresponding to a mean of 351.7 and a sample standard deviation of approximately 6.0. This modest variation provides initial evidence of stability in the evaluated configuration, while a larger number of runs and settings would be needed for broader conclusions.

\begin{table}[H]
    \centering
    \small
    \begin{tabular}{lc}
        \toprule
        Run & Unique extractions \\
        \midrule
        Seed 3 & 358 \\
        Seed 36 & 346 \\
        Seed 42 & 351 \\
        \midrule
        Mean $\pm$ sample std. & $351.7 \pm 6.0$ \\
        \bottomrule
    \end{tabular}
    \caption{CIFAR-100 extraction quantity across three random seeds.}
    \label{tab:seeds}
\end{table}

\section{Experiment Setup}
\label{app:setup}
\subsection{Setup for ADM and EDM}
For ADM, we utilize the released pre-trained checkpoints trained on \texttt{CIFAR10} 32 $\times$ 32, \texttt{CIFAR100} 32 $\times$ 32 and \texttt{Tiny-ImageNet 200} 32 $\times$ 32 \cite{kong2023efficient}. We perform the generation process using a 100-step DDIM sampler. We set guidance scale $s = 70.0$, temperature $\phi = 15.0$. As described in the algorithm \ref{alg:MGE_framework}, the diffusion process is split at $t_0=50$. In the first stage $t_0 \leq t \leq T$, we use a DDIM variance of $\sigma_0 = 1.0$, and in the second stage $0 \leq t \leq t_0$, we maintain $\sigma_1 = 1.0$ for consistency. We set the branch number $m = 4$. For conditional generation, the guidance subset is defined as $\mathcal{G} = \{4t : 5 \leq t \leq 16 \}$. This configuration ensures that the guidance is only applied during these specific steps, which helps balance the generation quality and the computational cost.

Similarly, for EDM, we follow the noise schedule and framework proposed by Karras et al.~\cite{karras2022elucidating}. We utilize the released pre-trained checkpoints trained on \texttt{CIFAR10} 32 $\times$ 32, \texttt{AFHQv2} 64 $\times$ 64 and \texttt{FFHQ} 64 $\times$ 64. We perform the generation process using a 25-step DDIM sampler. We set guidance scale $s = 20.0$, temperature $\phi = 15.0$. We apply the same guidance subset and parallel branch decoding settings as ADM. Since previous research has not yet investigated data extraction specifically for the \texttt{AFHQv2} and \texttt{FFHQ} datasets, and because the memorization behavior of models on these datasets is relatively weak, it is rare to see an $\ell\left(\hat{\boldsymbol{x}}, \boldsymbol{x} ; N_{\hat{\boldsymbol{x}}}\right) < 1.3$. Therefore, we set the filtering threshold to 1.3 and compare our results with the baseline under this same standard. This comparison allows us to demonstrate the effectiveness of our method in revealing training data leakage risks.

\subsection{Setup for Stable Diffusion Model}
For Stable Diffusion, we utilize the released pre-trained checkpoint Stable Diffusion v1.5 trained on \texttt{LAION-5B} \cite{rombach2022high}. We perform the generation process using a 25-step DDIM sampler. We set guidance scale $s = 70.0$, temperature $\phi = 10.0$. Unlike ADM, we apply guidance at different denoising stages. In experiments of Stable Diffusion, the guidance subset is defined as $\mathcal{G} = \{4t : 5 \leq t \leq 12 \}$ with the branch number $m = 4$. We also set $\sigma_0 = 1.0$ and $\sigma_1 = 1.0$ for parallel branch decoding to accelerate the sampling process.

\paragraph{Detection Method.} As clarified in Section \ref{sec:experiments}, we extract training data from \texttt{LAION-2B} for Stable Diffusion. As in \cite{carlini2023extracting}, we select the 500 prompts correspond to the most duplicate images in \texttt{LAION-2B} to generate images. Then, we use the membership inference with method high accuracy proposed in \cite{carlini2023extracting} to recognize the candidates. Moreover, to avoid some missing extractions, we compare all generated images to the original images correspond to the 500 prompts.  

\section{More Examples}
Next, we show more extracted data from various benchmark datasets by our proposed MGE. The results are in Figures \ref{fig:mem_cifar10}, \ref{fig:mem_cifar100}, and \ref{fig:mem_tini}. Furthermore, the results of EDM are shown in Figures \ref{fig:edm_mem_cifar10}, \ref{fig:edm_mem_afhqv2}, and \ref{fig:edm_mem_ffhq}. The results of Stable Diffusion v1.5 are shown in Figures \ref{fig:mem_sd}.

\clearpage

\onecolumn
\subsection{\texttt{CIFAR10} Extracted Images on ADM}

\begin{center}
    \includegraphics[width=\textwidth]{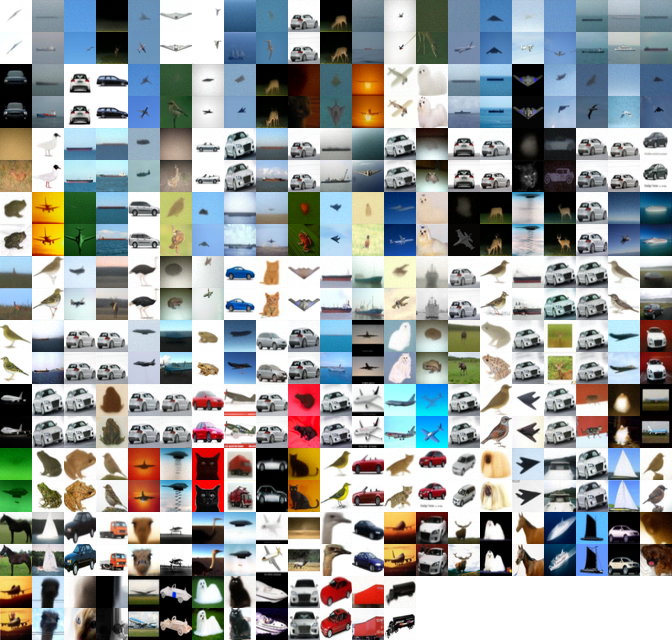}
    \captionof{figure}{Some images extracted by our MGE on \texttt{CIFAR10}.}
    \label{fig:mem_cifar10}
\end{center}

\clearpage

\subsection{\texttt{CIFAR100} Extracted Images on ADM}

\begin{center}
    \centering
    \includegraphics[width=\textwidth]{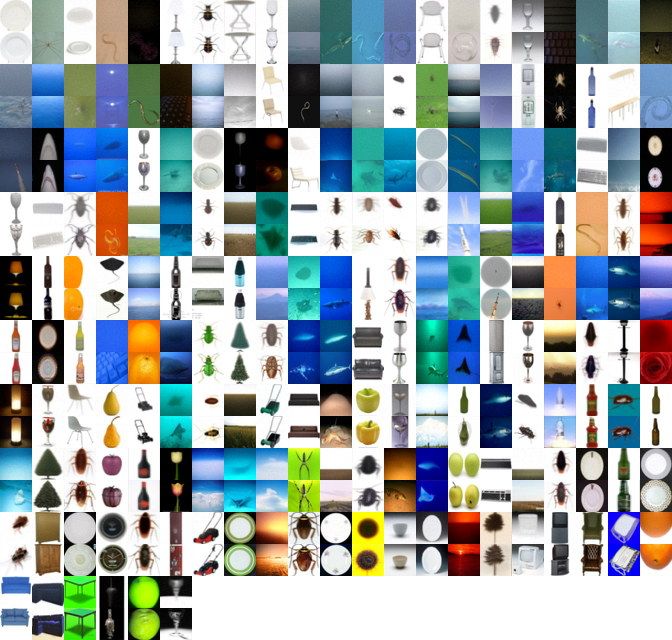}
    \captionof{figure}{Some images extracted by our MGE on \texttt{CIFAR100}.}
    \label{fig:mem_cifar100}
\end{center}

\clearpage

\subsection{\texttt{TINY-IN} Extracted Images on ADM}

\begin{center}
    \includegraphics[width=\textwidth]{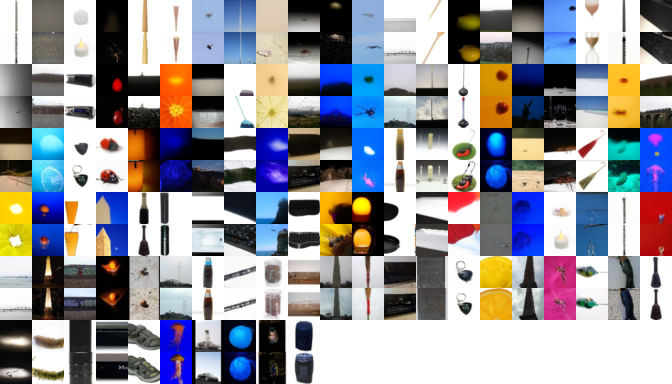}
    \captionof{figure}{Some images extracted by our MGE on \texttt{Tiny-IN}.}
    \label{fig:mem_tini}
\end{center}

\subsection{\texttt{CIFAR10} Extracted Images on EDM}

\begin{center}
    \centering
    \includegraphics[width=\textwidth]{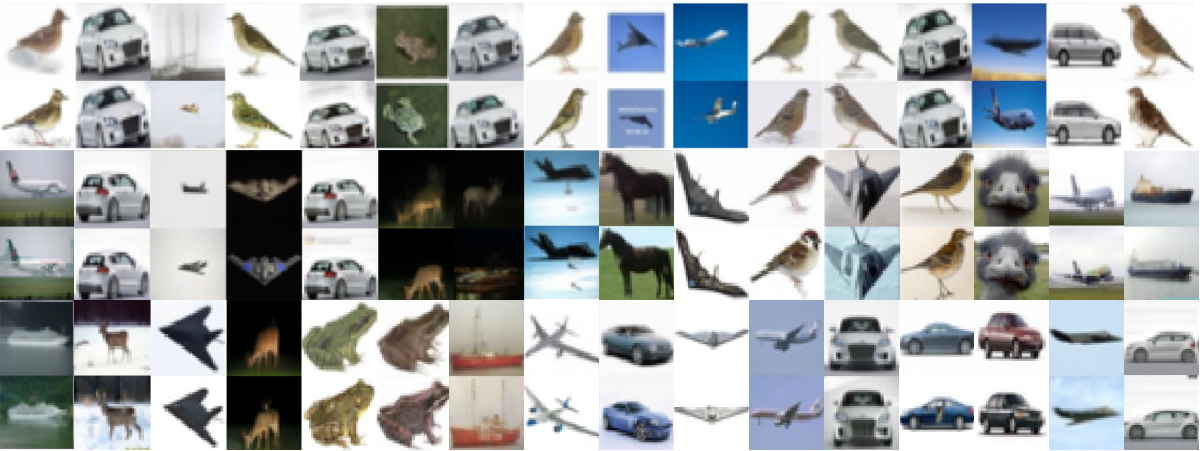}
    \captionof{figure}{Some images extracted by our MGE on \texttt{CIFAR10}.}
    \label{fig:edm_mem_cifar10}
\end{center}

\clearpage
\subsection{\texttt{AFHQv2} Extracted Images on EDM}

\begin{center}
    \includegraphics[width=0.75\textwidth]{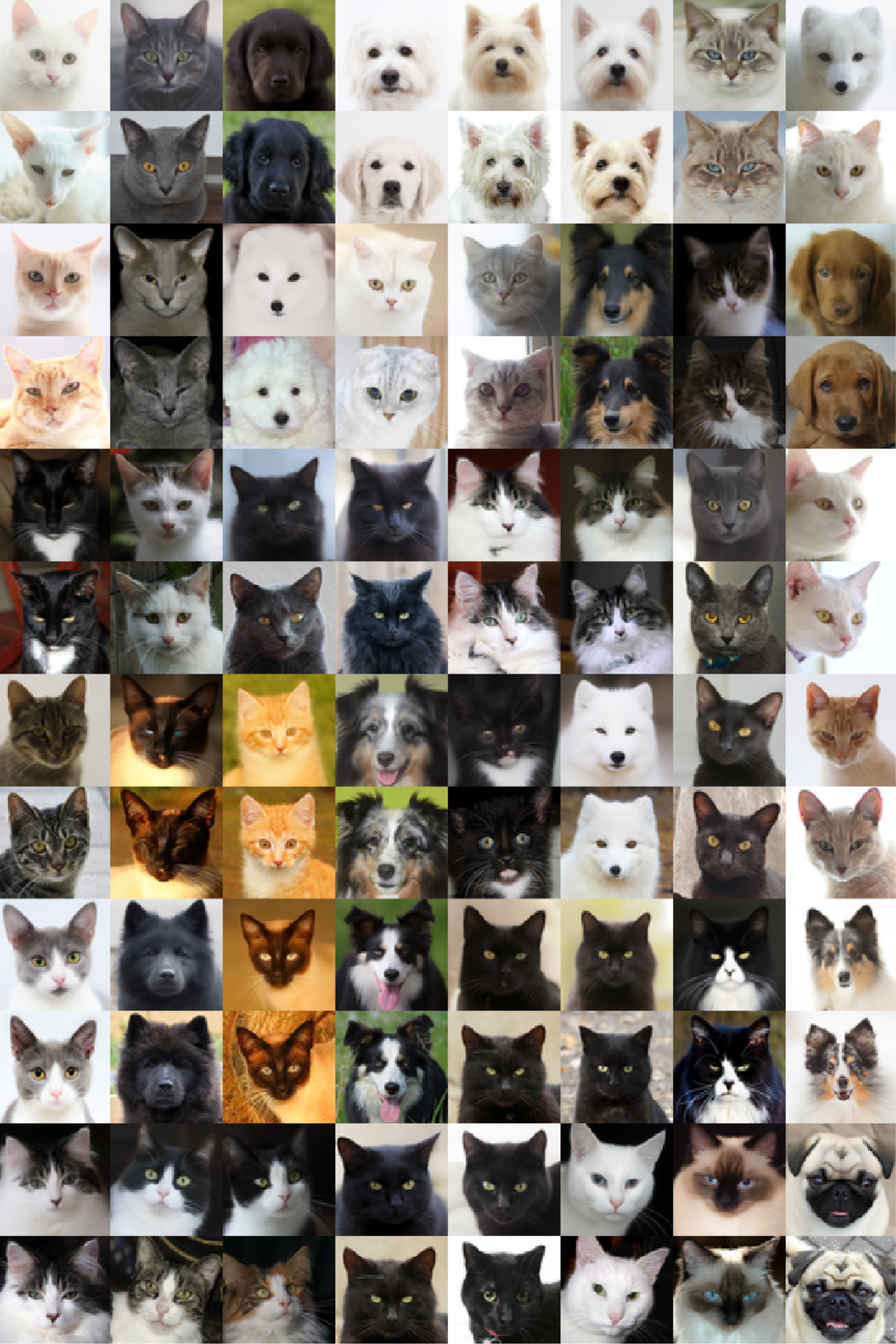}
    \captionof{figure}{Some images extracted by our MGE on \texttt{AFHQv2}.}
    \label{fig:edm_mem_afhqv2}
\end{center}

\clearpage
\subsection{\texttt{FFHQ} Extracted Images on EDM}

\begin{center}
    \includegraphics[width=0.75\textwidth]{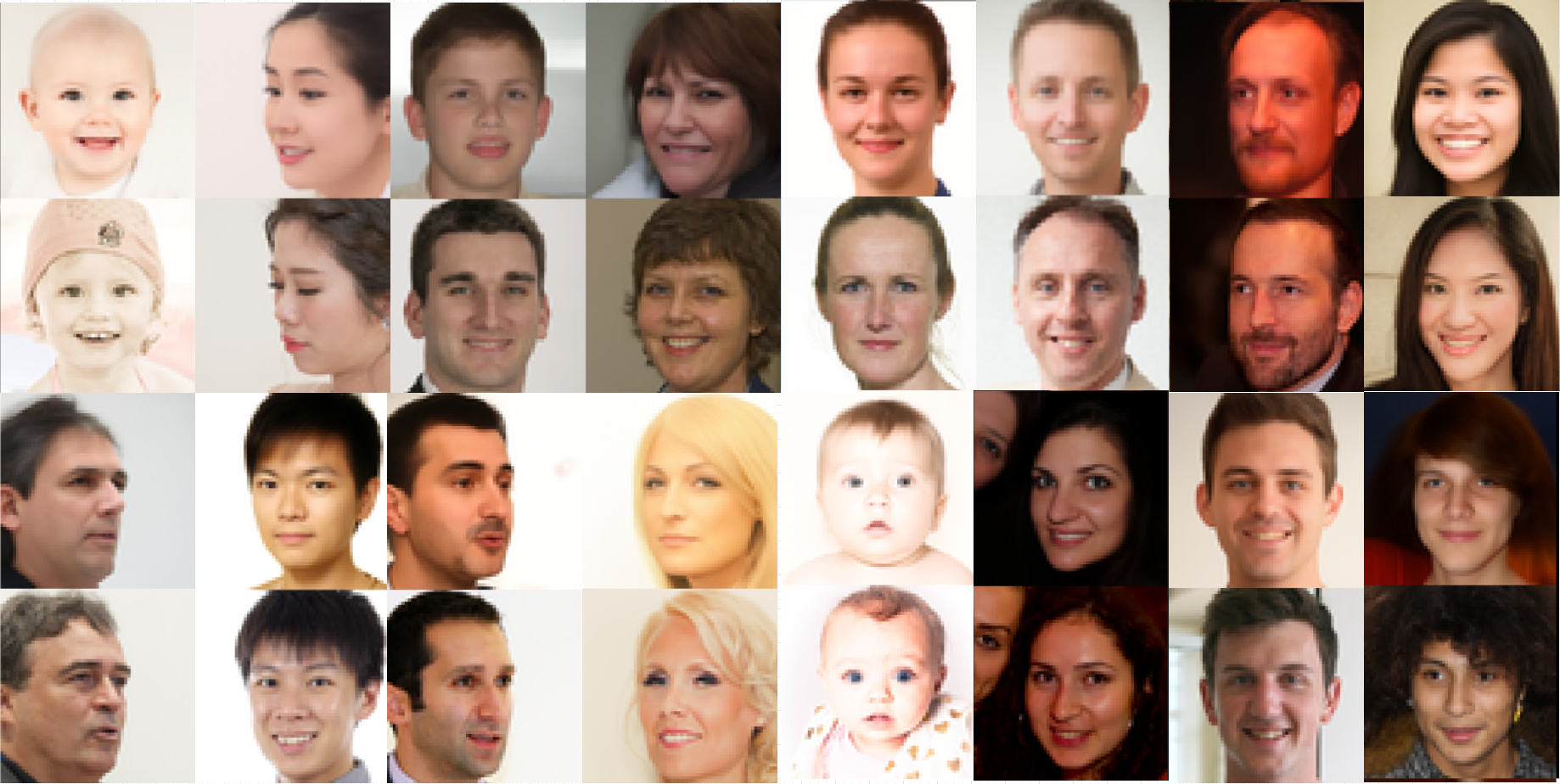}
    \captionof{figure}{Some images extracted by our MGE on \texttt{FFHQ}.}
    \label{fig:edm_mem_ffhq}
\end{center}

\subsection{Extracted Images on Stable Diffusion}

\begin{center}
    \includegraphics[width=0.8\textwidth]{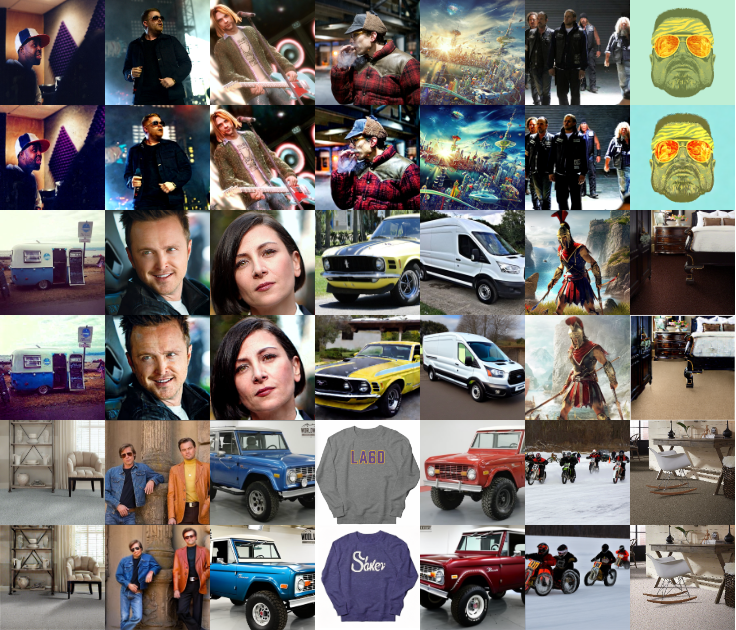}
    \captionof{figure}{Some images extracted by our MGE on Stable Diffusion v1.5.}
    \label{fig:mem_sd}
\end{center}

\end{document}